%% file: iclr2025_conference.tex
\documentclass{article} %
\usepackage{iclr2025_conference,times}

\input{math_commands.tex}

\usepackage{hyperref}
\usepackage{url}
\usepackage{array}
\usepackage{caption}
\usepackage{multirow}
\usepackage{ragged2e}
\usepackage{booktabs}
\usepackage{tabularray}
\usepackage{colortbl}
\usepackage{wasysym}
\usepackage{graphicx}
\usepackage{arydshln}
\usepackage{subcaption}
\usepackage{enumitem}
\usepackage{hhline}

\def\eg{{\em e.g.}}
\def\ie{{\em i.e.}}

\usepackage[capitalize,noabbrev]{cleveref}
\newtheorem{theorem}{Theorem}

\title{REEF: Representation Encoding Fingerprints for Large Language Models}

\author{\textbf{Jie Zhang}\textsuperscript{1,2{$\star$}},
\textbf{Dongrui Liu}\textsuperscript{1{$\star$}},
\textbf{Chen Qian}\textsuperscript{1,3},
\textbf{Linfeng Zhang}\textsuperscript{4},
\textbf{Yong Liu}\textsuperscript{3},
\textbf{Yu Qiao}\textsuperscript{1},
\textbf{Jing Shao}\textsuperscript{1}$^{\dag}$\\
$^1$ Shanghai Artificial Intelligence Laboratory \\
$^2$ University of Chinese Academy of Sciences \\
$^3$ Renmin University of China\\
$^4$ Shanghai Jiaotong University \\
\texttt{zhangjie@iie.ac.cn, qianchen2022@ruc.edu.cn}\\
\texttt{\{liudongrui,shaojing\}@pjlab.org.cn}
}

\iclrfinalcopy %
\begin{document}

\maketitle

\let\svthefootnote\thefootnote
\let\thefootnote\relax\footnotetext{$^\star$ Equal contribution\hspace{3pt} \hspace{5pt}$^{\dag}$ Corresponding author\hspace{5pt}}
\let\thefootnote\svthefootnote

\input{section/0_abstract}
\input{section/1_introduction}

\input{section/related_work}
\input{section/2_probe}

\input{section/3_cka}

\input{section/4_experiment}

\input{section/5_conclusion}

\subsubsection*{Reproducibility Statement}

To ensure the reproducibility of this study, we have uploaded the source code as part of the supplementary material. Furthermore, the code and datasets will be made available on GitHub after the completion of the double-blind review process, enabling others to replicate our study.

\bibliography{iclr2025_conference}
\bibliographystyle{iclr2025_conference}

\newpage
\appendix
\input{section/appendix}

\end{document}

%% file: math_commands.tex
\usepackage{amsmath,amsfonts,bm}

\def\eqref#1{equation~\ref{#1}}

\def\1{\bm{1}}

\DeclareMathAlphabet{\mathsfit}{\encodingdefault}{\sfdefault}{m}{sl}
\SetMathAlphabet{\mathsfit}{bold}{\encodingdefault}{\sfdefault}{bx}{n}

%% file: section/0_abstract.tex
\begin{abstract}
    Protecting the intellectual property of open-source Large Language Models (LLMs) is very important, because training LLMs costs extensive computational resources and data. Therefore, model owners and third parties need to identify whether a suspect model is a subsequent development of the victim model. To this end, we propose a training-free REEF to identify the relationship between the suspect and victim models from the perspective of LLMs' feature representations. Specifically, REEF computes and compares the centered kernel alignment similarity between the representations of a suspect model and a victim model on the same samples. This training-free REEF does not impair the model's general capabilities and is robust to sequential fine-tuning, pruning, model merging, and permutations. In this way, REEF provides a simple and effective way for third parties and models' owners to protect LLMs' intellectual property together. The code is available at https://github.com/tmylla/REEF.
\end{abstract}

%% file: section/1_introduction.tex
\section{Introduction}
\label{sec:intro}

The training process of Large Language Models (LLMs) requires extensive computational resources and time. Therefore, open-source models are usually released with specific licenses (\emph{e.g.}, Apache2.0, and LLaMA 2 Community License \citep{meta2023llama2license}) to protect their intellectual properties (IPs). 
Unfortunately, some developers claim to have trained their own LLMs but actually wrapped or fine-tuned based on other base LLMs (\eg, Llama-2 and MiniCPM-V) \citep{minicpmv_issue_196,huggingface_yi_34b_discussion}. It is urgent for model owners and third parties to identify \textit{whether the suspect model is a subsequent development of the victim model (e.g., Code-llama trained from Llama-2) or is developed from scratch (e.g., Mistral)}.

The key is to extract unique features (\ie, fingerprints) that can authenticate the victim model.
Watermarking methods artificially inject triggers into the victim model to make it generate specific content for identification \citep{DBLP:conf/acl/PengYWWZLJXSX23,xu2024instructional}.
However, watermarks introduce extra training costs and impair the model's general capabilities \citep{russinovich2024hey}, or even can be removed \citep{8682202,chen2023highfrequencymattersoverwritingattack}.
More crucially, these methods can not be applied to models that have already been open-released.
An alternative is to extract intrinsic features of the victim model, avoiding additional training and the compromise of capabilities.
Weight-based fingerprints are one of intrinsic features that allow calculating the similarity between a suspect model and a victim model's weights for identification \citep{zeng2023huref,refael2024slip}. 
However, these methods are fragile to major changes in weights, \eg, weight permutations, pruning, and extensive fine-tuning
\citep{fernandez2024functional,xu2024instructional}. 
This necessitates extracting more robust intrinsic features as fingerprints to identify victim models and protect their IPs.

\begin{figure}[tbp]
    \centering
    \includegraphics[width=\textwidth]{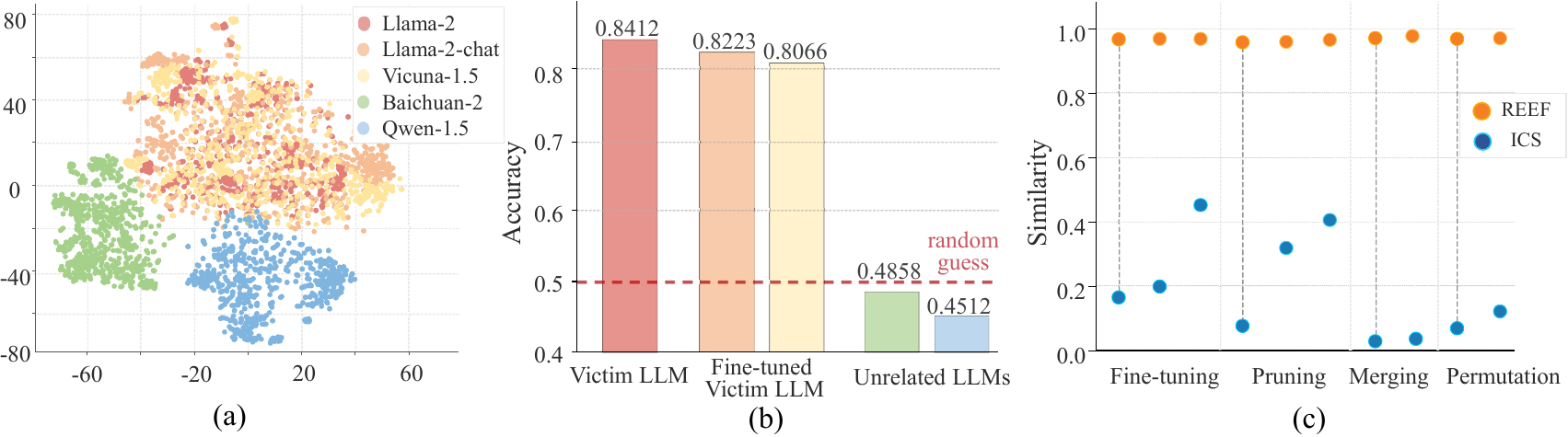} 
    \caption{(a) t-SNE visualization of different LLMs’ representations on the same samples. (b) Performance of classifiers trained on representations from the victim model evaluated on suspect models. (c) Robustness of REEF under variant LLMs that cause ICS \citep{zeng2023huref} ineffective.}
    \label{fig:vis}
\end{figure}

In this paper, we propose to solve this problem from the perspective of the \emph{feature representations} of LLMs, beginning with the following visualization analysis. It is generally acknowledged that
different models encode informative and intrinsic features based on their training data and model architecture, resulting in distinct feature representations across models \citep{mikolov2013linguistic,bolukbasi2016man,karras2021alias,chen2023surfacestatisticsscenerepresentations,zou2023representation}. 
Figure~\ref{fig:vis}(a) illustrates that the representations of Llama are markedly distinct from those of Baichuan and Qwen, while largely comparable to its fine-tuned models (\ie, Llama-chat and Chinese-llama).

Such findings inspire us to construct representation-based fingerprints.
Specifically, we apply neural networks 
to extract fingerprints of a victim model from its representation space. Figure~\ref{fig:vis}(b) shows that the classifier trained on representations of a victim model (\ie, Llama) can be generalized to its variant models (\eg, Llama-chat and Vicuna), but fail to other models (\eg, Baichuan and Qwen).
Although the effectiveness of representation-based fingerprints has been validated, such fingerprints still have limitations. 
On one hand, the input dimensions of neural networks are fixed, making them inapplicable to model pruning that alters the representation dimensions of the victim model \citep{frantar2023sparsegpt,xia2023flashllm,xia2024sheared}, which is prevalent in scenarios such as model compression for deployment on mobile devices.
On the other hand, these fingerprints lack robustness against representation permutations, a challenging issue because developers may intentionally manipulate model representations to evade detection \citep{zeng2023huref,refael2024slip}.

To this end, we propose a simple and effective approach, namely REEF, which is robust against pruning and evading detection. Specifically, REEF is a representation-based fingerprinting method that compares the Centered Kernel Alignment (CKA) similarity \citep{Kornblith2019} between the representations of the same samples from a suspect model and a victim model. Experimental results indicate that models derived from the victim model exhibit high similarity.
Moreover, REEF is resilient to dimensional changes, and we theoretically prove that CKA is invariant to column-wise permutations and scaling transformations.
Figure~\ref{fig:vis}(c) demonstrates that REEF maintains its effectiveness even under extreme conditions that cause weight-based methods \citep{zeng2023huref} ineffective. These conditions include extensive fine-tuning (using data with up to 700B tokens \citep{azerbayev2023llemma}), a high ratio pruning (up to 90\% of parameters \citep{ma2023llm}), model merging (merging LLMs with different architectures \citep{wan2024knowledge}), and permutations (parameter vector direction change through weight rearrangements \citep{fernandez2024functional}).

REEF utilizes the intrinsic feature from the perspective of representations to identify whether a suspect model is derived from a victim model. This training-free REEF does not impair model's general capabilities and is robust to various subsequent developments compared to weight-based fingerprints and watermarks. 
Consequently, REEF is a promising method for protecting the IPs of model owners and provides an efficient and effective way for third parties to review models, combating unethical or illegal activities such as unauthorized use or reproduction.

%% file: section/related_work.tex
\section{Related Work}
\label{sec:related_work}

Model fingerprinting protects IPs by allowing model owners and third parties to authenticate model ownership. There are two types of fingerprints for LLMs. One is injected fingerprints, which are artificially added during training or fine-tuning to facilitate model identification, such as watermarking methods \citep{DBLP:conf/acl/PengYWWZLJXSX23,xu2024instructional}. The other is intrinsic fingerprints, which are inherent properties that naturally emerge from the models' training data and architectures, including model weights (\textit{i.e.}, parameters) and feature representations, also known as embeddings or activations.

\paragraph{Injected Fingerprint.}
Watermarking methods inject a backdoor trigger into a victim model so that it will generate specific content in the presence of this trigger, allowing identification of whether a suspect model derives from the victim model. 
A large number of approaches embed the watermarks through backdoor attacks \citep{adi2018turning,zhang2018protecting,li2019prove}, and digital signature technology and hash functions \citep{guo2018watermarking,li2019piracy,zhu2020secure} are also used to design trigger words that contain the owner’s identity information to protect the IPs of DNNs. 
For generative language models, the high computational resource and time costs of training pose an urgent need to protect their IPs. Researchers have proposed various methods to inject watermarks as fingerprints to identify the victim model \citep{li2023plmmark,peng2023Watermark,kirchenbauer2023watermark,zhao2023protecting,russinovich2024hey,xu2024instructional}, but such methods inevitably impair the model's overall performance.

\paragraph{Intrinsic Fingerprint.}
Such fingerprints use the inherent and native attributes of the victim model, without requiring additional tuning which could impair the model's general capabilities, and are more stable and can not be removed. 
Model weights are one of the intrinsic features that can be used to compute the similarity of parameters between a suspect model and a victim model for identification \citep{zeng2023huref,refael2024slip}. 
Semantic analysis methods conduct statistical analysis on the content generated by different models, exploiting the linguistic patterns and semantic preferences exhibited by various LLMs as their unique fingerprints \citep{iourovitski2024hide,pasquini2024llmmap,mcgovern2024your}. 
However, both methods suffer from insufficient robustness \citep{xu2024instructional}. 
The internal representations of LLMs are derived from the data, strategies, and frameworks used during the training process, and serve as intrinsic features for model identification \citep{sevastjanova2022lmfingerprints}. For example, the logits space can be leveraged to identify the victim model \citep{yang2024fingerprint}. However, this approach remains highly sensitive to parameter permutation, posing significant challenges for effective fingerprinting.

%% file: section/2_probe.tex
\section{Exploring the Potential of Feature Representations as Fingerprints}
\label{sec:dnn}

In this section, we propose to utilize \emph{feature representations} as LLM fingerprints to identify whether a suspect model is a subsequent development of the victim model, based on the following two observations. (1) Feature representations of fine-tuned victim models are similar to feature representations of the original victim model, while the feature representations of unrelated models exhibit distinct distributions, as shown in Figure~\ref{fig:vis}(a). (2) Some high-level semantic concepts are ``linearly" encoded in the representation space of LLMs and can be easily classified, such as safety or unsafety and honest or dishonest \citep{zou2023representation,slobodkin-etal-2023-curious,qian2024towards}. 
According to these two observations, we can train a binary classifier on the representations of the victim model and then apply it to various suspect models' representations, \emph{i.e.,} LLMs derived from the victim model and unrelated LLMs. In this way, such a classifier may generalize to different fine-tuned victim models, because they have similar feature representations.

The binary classifier can employ various Deep Neural Network (DNN) architectures, such as a linear classifier, Multi-Layer Perceptron (MLP), Convolutional Neural Network (CNN), and Graph Convolutional Network (GCN). For training, we use the TruthfulQA dataset \citep{DBLP:conf/acl/LinHE22}, concatenating each question with its truthful answer as positive samples and with its false answer as negative samples. The dataset is split into training and test sets with a 4:1 ratio.
To evaluate the classifier's performance, we conduct experiments on LLMs of varying sizes. Specifically, we select Llama-2-7b and Llama-2-13b as the victim models, while derived models and unrelated LLMs serve as suspect models for comparison.

\begin{figure}[tbp]
    \centering
    \includegraphics[width=\textwidth]{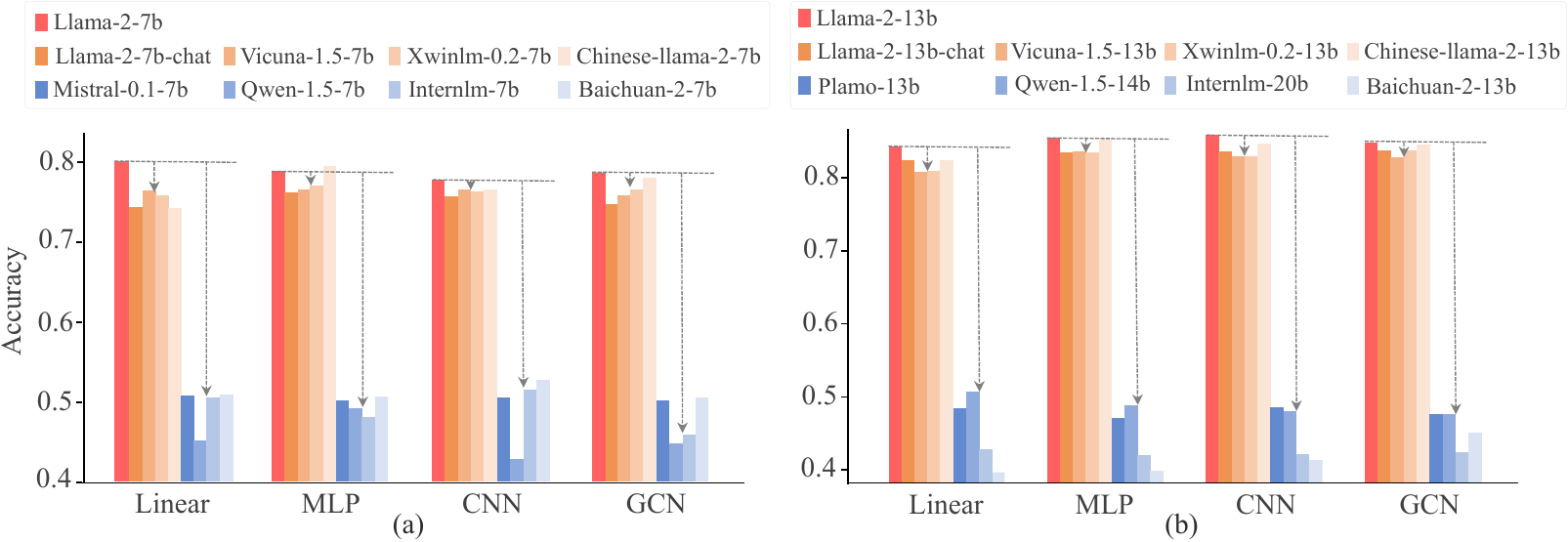} 
    \caption{Accuracies of classifiers trained on representations from the victim model: (a) Llama-2-7b as the victim model, (b) Llama-2-13b as the victim model.} 
    \label{fig:dnn}
\end{figure}

\paragraph{Classifiers trained on representations of a victim model can effectively generalize to its variants but not to others.}
Figure~\ref{fig:dnn}(a) shows that a classifier trained on the 18th layer representation of Llama-2-7b achieves approximately 80\% classification accuracy when applied to its fine-tuned models (\textit{e.g.}, Chinese-llama-2-7b). However, the accuracy drops to around 50\% on unrelated models (\textit{e.g.}, Mistral-0.1-7b), which is close to the level of random guessing. 
Classifiers trained on representations from other layers show the same results, as discussed in Appendix \ref{sec:dnn-appx}.
Additionally, similar findings are observed for Llama-2-13b (Figure~\ref{fig:dnn}(b)), indicating the scalability of the representation-based fingerprints. These experimental results indicate that representations can serve as fingerprints to protect the victim model’s IP.

\textbf{Challenges to using the classifier for victim model identification:} (1) DNNs have fixed input dimensions and cannot be applied to models pruned from the victim model, \eg, reducing representation dimensions. For example, the pruned models Sheared-llama-1.3b and Sheared-llama-2.7b have dimensions of 2048 and 2560, respectively \citep{xia2024sheared}. However, the classifier trained on Llama-2-7b can only process inputs of 4096 dimensions. (2) DNNs are not robust to permutations of the input feature representations, such as when columns are permuted through coupled matrix multiplications, which malicious developers might use to evade detection \citep{fernandez2024functional}.

%% file: section/3_cka.tex
\section{Robust Representation-based Fingerprinting with REEF}
\label{sec:cka}

To address the challenges of classifiers in victim model identification, we propose REEF, an advanced representation-based fingerprinting approach that can adapt to suspect models with varying representation dimensions and is robust to representation permutations.

REEF identifies whether a suspect model is derived from a victim model, given the representations of these two models on certain examples. Specifically, let $X \in \mathbb{R}^{m\times p_1}$ denote activations of the $l$-th layer from the suspect model on $m$ examples and $Y \in \mathbb{R}^{m\times p_2}$ denotes activations of the $l^{'}$-th layers from the victim model on the same $m$ examples, where $p_1$ is independent of $p_2$, meaning there is no limitation on dimensional consistency. Therefore, we need a similarity index $s(\cdot, \cdot)$ to measure representations' similarity between the suspect and victim models. In this way, a high $s(X,Y)$ score indicates that the suspect model is more likely derived from the victim model. In contrast, a low $s(X,Y)$ score means that the suspect model is less likely derived from the victim model.

\paragraph{Centered Kernel Alignment.} CKA \citep{Kornblith2019} is a similarity index based on Hilbert-Schmidt Independence Criterion (HSIC) \citep{gretton2005measuring}, which measures the independence between two sets of random variables. The CKA similarity between $X$ and $Y$ can be computed as follows
\begin{align}
    \text{CKA}(X, Y) = \frac{\text{HSIC}(X, Y)}{\sqrt{\text{HSIC}(X, X) \cdot \text{HSIC}(Y, Y)}},
\label{cka}
\end{align}
where $\text{HSIC}(X, Y) = \frac{1}{(m-1)^2}\text{tr}(K_XHK_YH)$. Specifically, $H = I - \frac{1}{m} \mathbf{1}\mathbf{1}^\text{T}$ is a centering matrix. $K_X$ and $K_Y$ are Gram matrices that measure the similarity of a pair
of examples based on kernel function $k$, \ie, $(K_X)_{ij} = k(X_i, X_j)$ and $(K_Y)_{ij} = k(Y_i, Y_j)$. $X_i$ and $X_j$ denote the $i$-th and $j$-th row of $X$, respectively. %

\paragraph{Kernel Selection.}
In this study, we consider a linear kernel and a Radial Basis Function (RBF) kernel. In the linear kernel case, Gram matrix $K_X = XX^{\top}$. In the RBF kernel case, $k(X_i, X_j) = \exp(-||X_i - X_j||_2^2/(2\sigma^2))$. Empirically, we discover that linear and RBF kernels obtain similar experimental results. Please see Section~\ref{subsec:validity} for more discussions. Unless otherwise specified, we adopt linear CKA due to its high efficiency.

\begin{theorem}
\label{theo:cka}
(Proof in Appendix \ref{ap:proof}) Given two matrices \( X \in \mathbb{R}^{m \times p_1} \) and \( Y \in \mathbb{R}^{m \times p_2} \), the CKA similarity score between \( X \) and \( Y \) is invariant under any permutation of the columns and column-wise scaling transformation. Formally, we have:
\begin{equation}
\label{eqn:cka}
\text{CKA}(X, Y) = \text{CKA}(XP_1, YP_2) = \text{CKA}(c_1X, c_2Y)
\end{equation}
where \( P_1 \in \mathbb{R}^{p_1 \times p_1} \) and \( P_2 \in \mathbb{R}^{p_2 \times p_2} \) denote permutation matrices. \( c_1 \in \mathbb{R}^{+} \) and \( c_2 \in \mathbb{R}^{+}\) are two positive scalars.
\end{theorem}

Theorem \ref{theo:cka} indicates that the CKA similarity score is theoretically invariant and robust to any column-wise permutations and scaling transformations. 
\cite{Kornblith2019} have shown that CKA is able to the correspondence between representations of different dimensions. Therefore, REEF is highly robust to various subsequent developments of the victim model in practical scenarios, including model pruning and representation permutation, ensuring accurate identification of the victim model through representation-based fingerprints to protect its IP.

%% file: section/4_experiment.tex
\section{Experiments}
\label{sec:experiments}

In this section, we provide a comprehensive evaluation of REEF. Section~\ref{subsec:validity} evaluates REEF's effectiveness in distinguishing LLMs derived from the victim model from unrelated models. Following this, Section~\ref{subsec:robust} assesses REEF's robustness to subsequent developments of the victim model, such as fine-tuning, pruning, merging, and permutations. Section~\ref{subsec:ablation} presents ablation studies on REEF across varying sample numbers and datasets. Finally, Section~\ref{subsec:discussion} discusses REEF's sensitivity to training data and its capacity for adversarial evasion.

\subsection{Effectiveness Verification}
\label{subsec:validity}

In this subsection, we demonstrate that REEF can effectively model the fingerprint from the representation. The CKA similarity between the victim model's representations and those of its derived models, as well as unrelated models, shows significant differences. This makes REEF a reliable fingerprinting method for protecting the victim model's IP.

\begin{figure}[tbp]
    \centering
    \includegraphics[width=\textwidth]{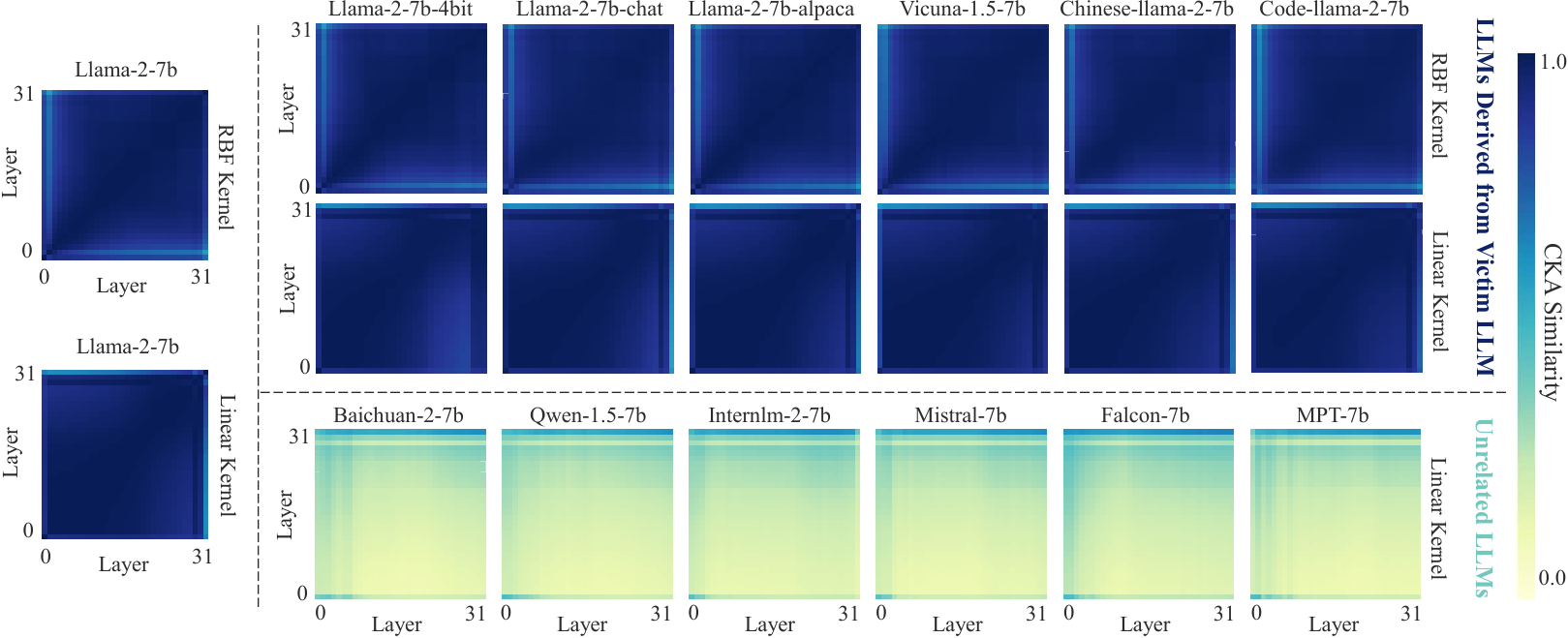} 
    \caption{Heatmaps depicting the CKA similarity between the representations of the victim LLM (Llama-2-7B) and those of various suspect LLMs on the same samples.}
    \label{fig:heatmap}
\end{figure}

\textbf{Settings.}
For the LLMs, we select Llama-2-7b as the victim model and choose a range of suspect models, including quantization and fine-tuned variants of Llama-2-7b (\textit{e.g.}, Llama-2-7b-chat, Code-llama-7b, and Llama-2-7b-4bit) as well as unrelated models (\textit{e.g.}, Qwen-1.5-7b, Baichuan-2-7b, and Mistral-7b). We use both a linear kernel and an RBF kernel to compute the layer-wise and inter-layer CKA similarity of representations between the victim and suspect models on 200 samples from the TruthfulQA dataset \citep{DBLP:conf/acl/LinHE22}. 
The resulting heatmap is shown in Figure~\ref{fig:heatmap}.

\textbf{REEF can accurately distinguish between models derived from the victim model and unrelated models.} 
As shown in Figure~\ref{fig:heatmap}, for LLMs derived from the victim model, the CKA similarity with the victim model is high, whereas unrelated LLMs show low similarity. This is reflected in the marked color contrast between the first two rows and the third row of the heatmap.
To quantify these results, the average similarity of LLMs derived from the victim model is 0.9585, which is substantially higher than that of unrelated LLMs, whose average similarity is 0.2361.
Additionally, for LLMs derived from the victim model, the similarity is notably high along the diagonal of the heatmaps, which represents the similarity between corresponding layers of the victim and suspect models, with an average of 0.9930. Furthermore, the inter-layer similarity is also significant, reaching 0.9707.

\textbf{Linear and RBF kernels yield similar results in identifying whether a suspect model is derived from the victim model.} 
As shown in the first two rows of Figure \ref{fig:heatmap}, the CKA similarity between the victim model and the LLMs derived from it, calculated using both the linear and RBF kernels, exceeded 0.95. This demonstrates that both kernels are suitable for fingerprinting in REEF. We adopt the linear CKA due to its higher computational efficiency.

\textbf{CKA from a single layer is sufficient for fingerprint identification.} 
The similarities between representations on a specific layer of the victim model and those of the derived and unrelated models differ significantly (\textit{e.g.}, 0.9973 and 0.2223 for layer 18, respectively).  Consequently, we focus on reporting the similarity at layer 18 in subsequent experiments, due to its informativeness and efficiency. The complete heatmap results are provided in Appendix~\ref{sec:heatmap}.

\subsection{Robustness Verification}
\label{subsec:robust}

In this subsection, we apply REEF to suspect models that are developed from a victim model through fine-tuning, pruning, merging, permutations, and scaling transformations. These techniques can introduce significant changes to the model's structure or parameters, making it challenging for existing methods to identify the victim model. However, REEF remains effective in these scenarios, demonstrating its robustness.

\subsubsection{Baseline Methods}

\textbf{Weight-based Fingerprinting Methods.} 
Following \cite{zeng2023huref}, we use model weight similarity methods, including PCS and ICS, to identify whether a suspect model is derived from a victim model. Specifically, PCS flattens all weight matrices and biases of an LLM into vectors and directly compares the cosine similarity between these vectors for the two models.
ICS constructs invariant terms from the weights of the last two layers and calculates the cosine similarity between these invariant terms for the two models.
A high cosine similarity indicates that the suspect model is derived from the victim model, and vice versa.

\textbf{Representation-based Fingerprinting Methods.} 
\cite{yang2024fingerprint}, referring to the Logits method, implements LLM fingerprinting by analyzing unique attributes of each LLM's logits output. This method evaluates the similarity between the output spaces of the victim and suspect models. A high similarity suggests that the suspect model is derived from the victim model. We conduct experiments on the TruthfulQA dataset to extract logit output for the suspect models.

\input{section/table-all}

\subsubsection{Fine-tuning}

\cite{xu2024instructional} point out that weight-based fingerprints are not reliable when models undergo extensive fine-tuning with larger deviations in parameters.
Given this challenge, we seek to assess the robustness of REEF under such demanding scenarios.

\textbf{Settings.}
We use Llama-2-7b as the victim model and select a diverse set of its fine-tuned models as suspect models, with fine-tuning (FT) data volumes ranging from 5 million to 700 billion tokens. The suspect models include Llama-2-finance-7b, Vicuna-1.5-7b, Wizardmath-7b, Chinese-llama-2-7b, Code-llama-7b, and Llemma-7b, with each model's fine-tuning data volume being 5M, 370M, 1.8B, 13B, 500B, and 700B tokens, respectively \citep{vicuna2023,luo2023wizardmath,Chinese-LLaMA-Alpaca,roziere2023code,azerbayev2023llemma}.

\textbf{REEF is robustness to extensive fine-tuning.}
As shown in Table~\ref{tab:reef}, even for models fine-tuned on datasets with up to 700B tokens (\ie, Llemma-7B), REEF still achieves a high similarity of 0.9962. In contrast, PCS becomes ineffective as early as fine-tuning with 1.8B tokens (\ie, Wizardmath-7b). ICS performance significantly degrades with increasing fine-tuning data volume, with 13B tokens (\ie, Chinese-llama-2-7b) and 500B tokens (\ie, Code-llama-7B) yielding similarity of 0.4996 and 0.2550, respectively. Although the Logits method shows relatively less degradation, it still exhibits sensitivity to the volume of fine-tuning data. Notably, Logits method is particularly sensitive to changes in the vocabulary, \eg, Chinese-llama-2-7b has expanded its vocabulary during fine-tuning, yielding a lower similarity than Code-llama-7b (0.7033 vs 0.7833), despite being fine-tuned on a smaller dataset (13B vs 500B tokens).

\textbf{Discussion about how much fine-tuning data could make REEF ineffective.}
Despite fine-tuning Llama-2-7b to Llemma-7b with 700B tokens \citep{azerbayev2023llemma}, the fine-tuning data is one-third of Llama-2-7b's 2T token pre-training data, yet REEF remains effective. We question whether REEF would remain effective with continued increases in fine-tuning data.
Before delving into this discussion, two statements are listed: (1) To the best of our know, Llemma-7b is the most extensively fine-tuned Llama-2-7b model, nearly 700B tokens for fine-tuning, and REEF has shown robustness in this context; (2) Code-llama-7b \citep{roziere2023code} reports that fine-tuning with 500B tokens requires 4.4T of disk size and 25,000 GPU hours, fine-tuning on this scale is costly.
Such a considerable cost limits further extensive fine-tuning. REEF appears effective in current fine-tuning scenarios.

\subsubsection{Model Pruning}

Pruning is widely used in model compression for edge deployment, \eg, serving for mobile devices and autonomous driving \citep{vadera2021methodspruningdeepneural,wang2024modelcompressionefficientinference, lin2024mlp}. However, pruning could significantly alter both the structural integrity and representation dimensions of models \citep{ma2023llm,frantar2023sparsegpt,zhu2023survey}, posing challenges for fingerprint identification. To this end, we test REEF on various pruned models of the victim model Llama-2-7b.

\textbf{Settings.}
We use Llama-2-7b as the victim model and various pruned models of it as suspect models. First, we select several pruned models using different pruning strategies, including structured pruning (\eg Sheared-llama \citep{xia2024sheared}), and unstructured pruning (\eg, SparseGPT \citep{frantar2023sparsegpt}, GBLM-Pruner \citep{das2023beyond}, and Wanda \citep{sun2023wanda}). These methods prune the models at specific ratios, followed by post-training (\textit{e.g.}, continued pre-training or instruction-tuning) to ensure the pruned models maintain their capabilities. Second, we apply LLM-Pruner \citep{ma2023llm} to prune Llama-2-7b into smaller suspect models at arbitrary pruning ratios, without post-training. For example, we apply block pruning to reduce Llama-2-7b's parameters by 10\% to as much as 90\%, and layer pruning to reduce the number of layers by 3 to as much as 27.

\textbf{REEF is robust to various pruning strategies.}
As shown in Table~\ref{tab:reef}, for structured pruned models, REEF consistently achieves accurate fingerprint identification across all Sheared-llama models, with similarities exceeding 0.9278. In contrast, PCS fails in this scenario, consistently yielding a similarity score of zero. ICS does not perform well, \eg, the similarity for the 1.3B pruned model drops to 0.3512. The Logits method, which relies on the output space, remains unaffected unless the pruning alters the logits themselves.
For unstructured pruned models, all methods are capable of identifying the victim model, with all similarities exceeding 0.94. In summary, REEF and the Logits method remain robust across all pruned models.

\begin{figure}[t]
    \centering
    \includegraphics[width=\textwidth]{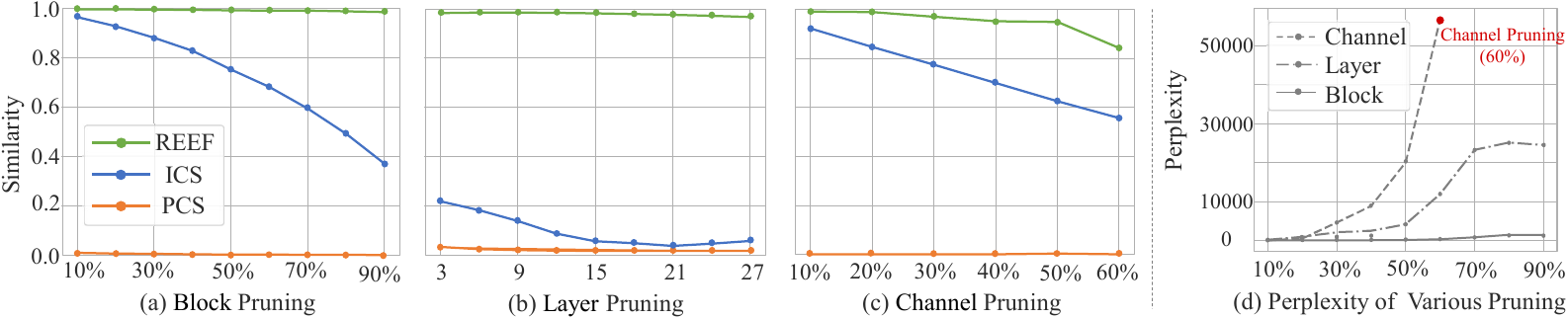} 
    \caption{(a)-(c) Similarity between pruned models and the victim model across three pruning strategies at various pruning ratios. (d) Perplexity of the three pruning strategies.}
    \label{fig:pruning}
\end{figure}

\textbf{REEF is robustness to pruning ratio, even up to 90\%.} 
Figure~\ref{fig:pruning} shows that REEF remains effective even with significant pruning, including block pruning of up to 90\% of parameters, layer pruning of up to 27 layers, and channel pruning of up to 60\%. Figure~\ref{fig:pruning}(d) illustrates that perplexities are particularly high in these scenarios, especially with 60\% channel pruning. As noted by \cite{ma2023llm}, channel pruning affects all layers, but the first and last layers are critical for maintaining model integrity, thus pruning is limited to 60\%. 
In contrast, PCS fails in all pruning scenarios, and ICS's effectiveness diminishes as the pruning ratio increases, ultimately failing under layer pruning. These findings highlight REEF as the most robust and reliable method for fingerprint identification across various pruning ratios.

\subsubsection{Model Merging}

Model merging is an effective technique that merges multiple separate models with different capabilities to build a universal model without needing access to the original training data or expensive computation \citep{yang2024modelmergingllmsmllms}.
Differing from other sections, the merged model is derived from several victim models, which pose a challenge in identifying all of them. In this subsection, we study two types of model merging: weight-based and distribution-based.

\textbf{Settings.}
For weight merging, we select Evollm-jp-7b \citep{akiba2024evolutionary} as the suspect model, which merges three victim models with the same architecture (\ie, Shisa-gamma-7b-v1, Wizardmath-7b-1.1, and Abel-7b-002) by weighted parameters. For distribution merging, we choose Fusellm-7b \citep{wan2024knowledge} and Fusechat \citep{wan2024fusechat} as suspect models, respectively. Fusellm-7b merges three victim LLMs with distinct architectures but with same scale: Llama-2-7b, Openllama-2-7b, and Mpt-7b. Fusechat merges several chat LLMs of varied architectures and scales, we investigate Internlm2-chat-20b, Mixtral-8x7b-instruct, and Qwen-1.5-chat-72b as suspect models.

\textbf{REEF is robust across both weight and distribution merging scenarios.}
For weight merging, REEF consistently achieves high accuracy in identifying the origins of merged models, with similarities ranging from 0.9526 to 0.9996, as shown in Table~\ref{tab:reef}. ICS, PCS, and the Logits method also perform well in this scenario.
For distribution merging at the same scales (\ie, Fusellm-7b), REEF continues to perform well, accurately identifying the victim model Llama-2-7b with a similarity of 0.9996. Additionally, it remains effective for Openllama-2-7b and Mpt-7b, with similarities of 0.6713 and 0.62, respectively. However, ICS struggles significantly in this scenario, with all three original victim models achieving low similarities. Although PCS and the Logits method can identify Llama-2-7b, their performance drops sharply for Openllama-2-7b and Mpt-7b, with similarities of nearly 0. 
For distribution merging at the different scales (\ie, Fusechat-7b), REEF is the only method that continues to work for identifying victim models, while the other methods fail, demonstrating its consistent reliability in this scenario.
Based on these findings, REEF is robust across various merging strategies and can identify all victim models for the merged model.

\subsubsection{Permutation and Scaling Transformation} 

There are approaches that could camouflage the model without changing its architecture or affecting its output \citep{zeng2023huref}. Malicious developers may modify the model by employing dimension permutation or coupled matrix multiplications to evade some fingerprint detection methods \citep{fernandez2024functional}. This section aims to experiment with the robustness of various fingerprinting methods in addressing this type of evasion.

\textbf{Settings.}
We select Llama-2-7b, Mistral-7b, and Qwen-1.5-7b as victim models, applying column-wise permutations or scaling transformations (with a scaling factor of 0.8) to both their weight matrices and feature representations. These operations simulate evasion techniques that malicious developers might use, enabling us to compare the similarities of the weights and representations before and after the operations.

\textbf{REEF is invariant and robust to any column-wise permutations and scaling transformations, as proved by the Theorem~\ref{theo:cka}.}
As shown in Table~\ref{tab:reef}, the CKA similarity computed by REEF remains consistently at 1 before and after the permutation or scaling transformations, indicating that REEF is invariant to these operations and robust against evasion techniques. However, other methods such as ICS, PCS, and Logits, while robust to scaling transformations, exhibit a significant drop in similarity under permutation, with values nearly dropping to 0. These results further reinforce that REEF is a highly reliable fingerprinting method in practical applications against malicious developers.

\subsection{Ablation Study}
\label{subsec:ablation}

\begin{figure}[t]
    \centering
    \includegraphics[width=\textwidth]{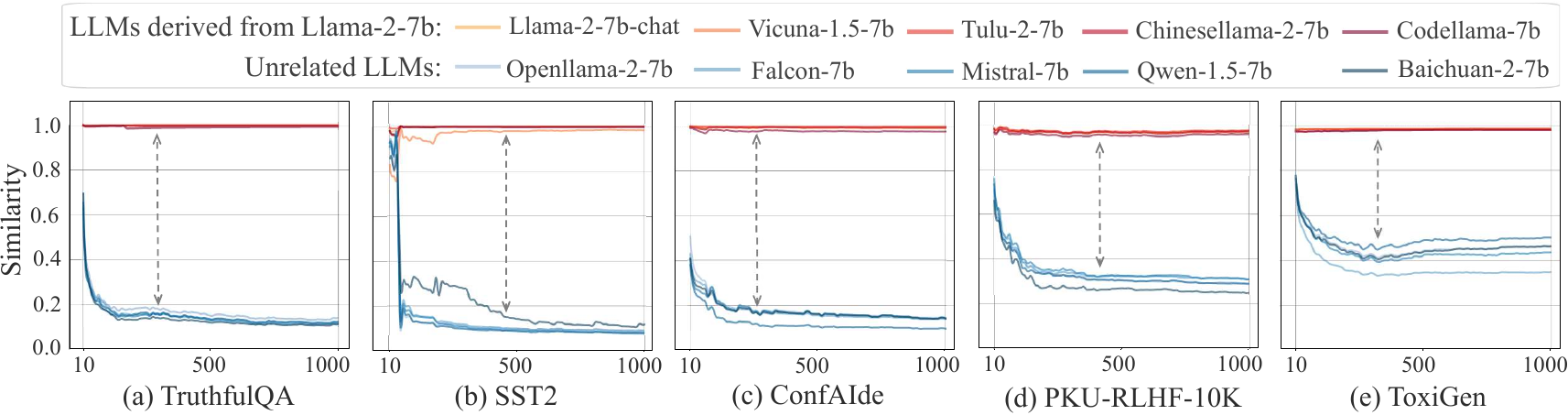} 
    \caption{Illustration of the CKA similarity between the representations of the victim LLM (Llama-2-7B) and various suspect LLMs across different datasets as sample number increases.}
    \label{fig:ablation}
\end{figure}

\textbf{\textit{Number of Samples} } 
To evaluate the impact of sample number on the performance of REEF, we conduct an ablation study using samples from TruthfulQA, ranging from 10 to 1000 in intervals of 10. We use Llama-2-7b as the victim model and select 10 suspect models, consisting of 5 LLMs derived from Llama-2-7b and 5 unrelated LLMs. 
We then calculate the CKA similarity between the sample representations of each suspect model and those of Llama-2-7b at different sample numbers. Figure~\ref{fig:ablation}(a) illustrates the similarities for various models as the number of samples increases.

\textbf{REEF is highly efficient regarding the number of samples required for robust model fingerprinting.}
Figure~\ref{fig:ablation}(a) shows that the similarities for most models stabilize after 200-300 samples, suggesting that REEF can achieve reliable fingerprint identification with a smaller sample number. Notably, LLMs derived from Llama-2-7b (\eg, Chinese-lama-2-7b and Code-llama-7b) consistently maintain high similarities close to 1.0 across all sample numbers. This indicates that these models potentially share the same representation space as the victim model, verifying that representation is an intrinsic feature for fingerprinting.
In contrast, unrelated LLMs (\eg, Qwen-7b-v1.5 and Mistral-7b) exhibit lower similarities that gradually decrease and stabilize at levels below 0.2 as the number of samples increases. This suggests that these models are more distinct and require a larger number of samples for accurate fingerprinting. 
Overall, few samples from TruthfulQA are effective for REEF in identifying LLMs derived from the victim model compared to unrelated LLMs.

\textbf{\textit{Different Datasets} }
To assess the effectiveness of REEF across various data types, we also conduct experiments using SST2 \citep{socher2013recursive}, ConfAIde \citep{confaide2023}, PKU-SafeRLHF \citep{ji2024pku}, and ToxiGen \citep{hartvigsen2022toxigen}. Following the same settings described in the previous section, we plot the similarities between the victim model and various suspect models for different datasets as the number of samples increases, as shown in Figure~\ref{fig:ablation}(b)-(e).

\textbf{REEF is effective across various datasets.}
Figure~\ref{fig:ablation}(b)-(e) show that the similarity between the victim model and its derived LLMs is significantly higher than the similarity with unrelated LLMs across different datasets. This clear distinction demonstrates that REEF can effectively identify whether a suspect model is derived from the victim model.
Furthermore, the gap in the similarity between derived LLMs and unrelated LLMs varies by dataset, \eg, the gap is approximately 0.8 on TruthfulQA and about 0.5 on ToxiGen. A larger gap indicates a stronger identification capability. Our findings suggest that while REEF is effective across diverse datasets, TruthfulQA emerges as the optimal choice for model fingerprinting, as it exhibits the most substantial differentiation in similarity between LLMs derived from the victim model and unrelated LLMs.

\subsection{Further Discussion}
\label{subsec:discussion}

\begin{figure}[t]
    \centering
    \includegraphics[width=0.96\textwidth]{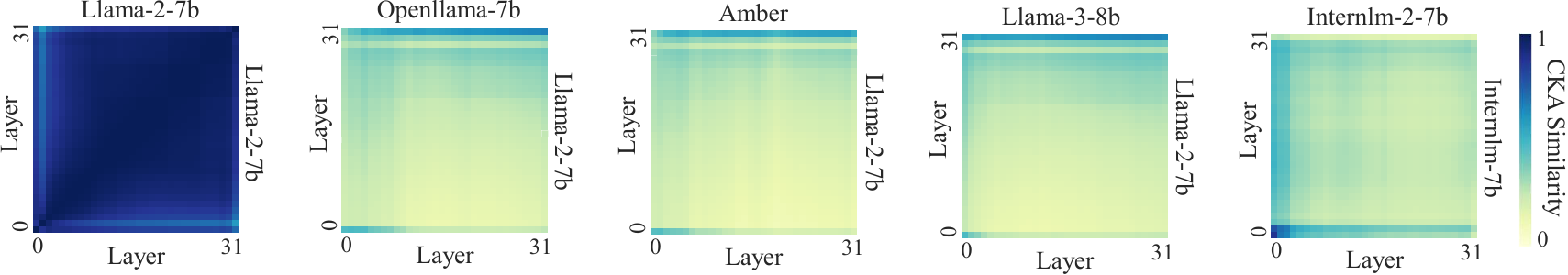} 
    \caption{Heatmaps depicting the CKA similarity between the representations of paired LLMs with the same architecture but different pre-training data.}
    \label{fig:discussion}
\end{figure}

\textbf{REEF can distinguish between models with the same architecture but different pre-training data.} 
Openllama-7b \citep{openlm2023openllama} and Amber \citep{liu2023llm360} are open-source LLMs that utilize the same Llama architecture but are trained from scratch on distinct pre-training datasets.
Figure~\ref{fig:discussion} demonstrates that REEF effectively identifies the differences between Llama-2-7b and both Openllama-7b and Amber, as evidenced by their low similarities.
Additionally, similar results are observed when analyzing LLMs across different generations, such as Llama-2-7b versus Llama-3-8b, and Internlm-7b versus Internlm2-7b. Each of these models reflects variations in pre-training data and strategies, which REEF accurately identifies.

\textbf{Malicious developers fail to fine-tune models with a customized loss function to evade detection by the REEF.} 
We assume these developers are aware of the REEF approach and attempt to design customized loss functions during fine-tuning to bypass detection. Since REEF relies on the observation that developed LLMs share similar representational spaces with the victim model. The developer may use the customized loss function to widen the gap between the two representations. Experimental results in Appendix~\ref{sec:fine-tuning} indicate that such fine-tuning seriously damage the model's general capabilities and renders the fine-tuned models unusable. This is because the capabilities of LLMs stem from their representational distribution, and such intentional fine-tuning inevitably leads to the model losing its language modeling ability. Therefore, malicious developers are unable to evade REEF detection through this method.

%% file: section/table-all.tex
\begin{table}[tb]
\centering
\caption{Similarity of various LLM fingerprinting methods applied to suspect models developed through fine-tuning, pruning, merging, permutations, and scaling transformations. 
In this table, \colorbox[rgb]{0.925,1,0.925}{\phantom{xx}} indicate similarity greater than 0.8, \colorbox[rgb]{0.925,0.925,1}{\phantom{xx}} indicate similarity between 0.5 and 0.8, and \colorbox[rgb]{1,0.925,0.925}{\phantom{xx}} indicate similarity less than 0.5.}
\label{tab:reef}
\scalebox{0.74}{
\renewcommand{\arraystretch}{1.}
\begin{tabular}{ccccccc} 
\hline\hline
                       & \multicolumn{6}{c}{{\cellcolor[rgb]{0.918,0.918,0.918}}\textbf{Model Fine-tuning }}                                                                                                                                                                                                                                                                                                                                                              \\
                       & \begin{tabular}[c]{@{}c@{}}Llama-2-finance-7b\\(5M Tokens)\end{tabular} & \begin{tabular}[c]{@{}c@{}}Vicuna-1.5-7b\\(370M Tokens)\end{tabular} & \begin{tabular}[c]{@{}c@{}}Wizardmath-7b\\(1.8B Tokens)\end{tabular}  & \begin{tabular}[c]{@{}c@{}}Chinesellama-2-7b\\(13B Tokens)\end{tabular} & \begin{tabular}[c]{@{}c@{}}Codellama-7b\\(500B Tokens)\end{tabular} & \begin{tabular}[c]{@{}c@{}}Llemma-7b\\(700B Tokens)\end{tabular}       \\ 
\hline
\textbf{PCS}           & {\cellcolor[rgb]{0.894,1,0.894}}0.9979                                  & {\cellcolor[rgb]{0.894,1,0.894}}0.9985                                & {\cellcolor[rgb]{1,0.894,0.894}}0.0250                                & {\cellcolor[rgb]{1,0.894,0.894}}0.0127                                  & {\cellcolor[rgb]{1,0.894,0.894}}0.0105                              & {\cellcolor[rgb]{1,0.894,0.894}}0.0098                                 \\
\textbf{ICS}           & {\cellcolor[rgb]{0.894,1,0.894}}0.9952                                  & {\cellcolor[rgb]{0.894,1,0.894}}0.9949                                & {\cellcolor[rgb]{0.894,1,0.894}}0.9994                                & {\cellcolor[rgb]{1,0.894,0.894}}0.4996                                  & {\cellcolor[rgb]{1,0.894,0.894}}0.2550                              & {\cellcolor[rgb]{1,0.894,0.894}}0.2257                                 \\
\textbf{Logits}        & {\cellcolor[rgb]{0.894,1,0.894}}0.9999                                  & {\cellcolor[rgb]{0.894,1,0.894}}0.9999                                & {\cellcolor[rgb]{0.894,1,0.894}}0.9999                                & {\cellcolor[rgb]{0.894,0.894,1}}0.7033                                  & {\cellcolor[rgb]{0.894,0.894,1}}0.7833                              & {\cellcolor[rgb]{0.894,0.894,1}}0.6367                                 \\
\textbf{REEF}          & {\cellcolor[rgb]{0.894,1,0.894}}0.9950                                  & {\cellcolor[rgb]{0.894,1,0.894}}0.9985                                & {\cellcolor[rgb]{0.894,1,0.894}}0.9979                                & {\cellcolor[rgb]{0.894,1,0.894}}0.9974                                  & {\cellcolor[rgb]{0.894,1,0.894}}0.9947                              & {\cellcolor[rgb]{0.894,1,0.894}}0.9962                                 \\ 
\hline\hline
                       & \multicolumn{6}{c}{{\cellcolor[rgb]{0.918,0.918,0.918}}\textbf{Structured~}~\textbf{Pruning }}                                                                                                                                                                                                                                                                                                                                                   \\
                       & \begin{tabular}[c]{@{}c@{}}Sheared-llama-\\1.3b-pruned\end{tabular}     & \begin{tabular}[c]{@{}c@{}}Sheared-llama-\\1.3b\end{tabular}          & \begin{tabular}[c]{@{}c@{}}Sheared-llama-\\1.3b-sharegpt\end{tabular} & \begin{tabular}[c]{@{}c@{}}Sheared-llama-\\2.7b-pruned\end{tabular}     & \begin{tabular}[c]{@{}c@{}}Sheared-llama-\\2.7b\end{tabular}        & \begin{tabular}[c]{@{}c@{}}Sheared-llama-\\2.7b-sharegpt\end{tabular}  \\ 
\hline
\textbf{PCS}           & {\cellcolor[rgb]{1,0.894,0.894}}0.0000                                  & {\cellcolor[rgb]{1,0.894,0.894}}0.0000                                & {\cellcolor[rgb]{1,0.894,0.894}}0.0000                                & {\cellcolor[rgb]{1,0.894,0.894}}0.0000                                  & {\cellcolor[rgb]{1,0.894,0.894}}0.0000                              & {\cellcolor[rgb]{1,0.894,0.894}}0.0000                                 \\
\textbf{ICS}           & {\cellcolor[rgb]{1,0.894,0.894}}0.4927                                  & {\cellcolor[rgb]{1,0.894,0.894}}0.3512                                & {\cellcolor[rgb]{1,0.894,0.894}}0.3510                                & {\cellcolor[rgb]{0.894,0.894,1}}0.6055                                  & {\cellcolor[rgb]{1,0.894,0.894}}0.4580                              & {\cellcolor[rgb]{1,0.894,0.894}}0.4548                                 \\
\textbf{Logits}        & {\cellcolor[rgb]{0.894,1,0.894}}0.9967                                  & {\cellcolor[rgb]{0.894,1,0.894}}0.9999                                & {\cellcolor[rgb]{0.894,1,0.894}}0.9999                                & {\cellcolor[rgb]{0.894,1,0.894}}0.9967                                  & {\cellcolor[rgb]{0.894,1,0.894}}0.9999                              & {\cellcolor[rgb]{0.894,1,0.894}}0.9999                                 \\
\textbf{\textbf{REEF}} & {\cellcolor[rgb]{0.894,1,0.894}}0.9368                                  & {\cellcolor[rgb]{0.894,1,0.894}}0.9676                                & {\cellcolor[rgb]{0.894,1,0.894}}0.9710                                & {\cellcolor[rgb]{0.894,1,0.894}}0.9278                                  & {\cellcolor[rgb]{0.894,1,0.894}}0.9701                              & {\cellcolor[rgb]{0.894,1,0.894}}0.9991                                 \\
\hline\hline
                       & \multicolumn{3}{c}{{\cellcolor[rgb]{0.918,0.918,0.918}}\textbf{Unstructured~~Pruning}}                                                                                                                                  & \multicolumn{3}{c}{{\cellcolor[rgb]{0.918,0.918,0.918}}\textbf{Distribution Merging (Fusechat-7b)}}                                                                                                                                           \\
                       & \begin{tabular}[c]{@{}c@{}}Sparse-llama-\\2-7b\end{tabular}             & \begin{tabular}[c]{@{}c@{}}Wanda-llama-\\2-7b\end{tabular}            & \begin{tabular}[c]{@{}c@{}}GBLM-llama-\\2-7b\end{tabular}             & \begin{tabular}[c]{@{}c@{}}Internlm2-chat-\\20b\end{tabular}            & \begin{tabular}[c]{@{}c@{}}Mixtral-8x7b-\\instruct\end{tabular}     & \begin{tabular}[c]{@{}c@{}}Qwen-1.5-chat-\\72b\end{tabular}            \\
\hline
\textbf{PCS}           & {\cellcolor[rgb]{0.894,1,0.894}}0.9560                                  & {\cellcolor[rgb]{0.894,1,0.894}}0.9620                                & {\cellcolor[rgb]{0.894,1,0.894}}0.9616                                & {\cellcolor[rgb]{1,0.894,0.894}}0.0000                                  & {\cellcolor[rgb]{1,0.894,0.894}}0.0000                              & {\cellcolor[rgb]{1,0.894,0.894}}0.0000                                 \\
\textbf{ICS}           & {\cellcolor[rgb]{0.894,1,0.894}}0.9468                                  & {\cellcolor[rgb]{0.894,1,0.894}}0.9468                                & {\cellcolor[rgb]{0.894,1,0.894}}0.9478                                &     {\cellcolor[rgb]{1,0.894,0.894}}0.1772     &     {\cellcolor[rgb]{1,0.894,0.894}}0.0105         &     {\cellcolor[rgb]{1,0.894,0.894}}0.0635                                                                   \\
\textbf{Logits}        & {\cellcolor[rgb]{0.894,1,0.894}}0.9999                                  & {\cellcolor[rgb]{0.894,1,0.894}}0.9999                                & {\cellcolor[rgb]{0.894,1,0.894}}0.9999                                & {\cellcolor[rgb]{1,0.894,0.894}}0.0000                                  & {\cellcolor[rgb]{1,0.894,0.894}}0.0000                              & {\cellcolor[rgb]{1,0.894,0.894}}0.0000                                 \\
\textbf{REEF}          & {\cellcolor[rgb]{0.894,1,0.894}}0.9985                                  & {\cellcolor[rgb]{0.894,1,0.894}}0.9986                                & {\cellcolor[rgb]{0.894,1,0.894}}0.9991                                & {\cellcolor[rgb]{0.894,1,0.894}}0.9278                                  & {\cellcolor[rgb]{0.894,1,0.894}}0.9701                              & {\cellcolor[rgb]{0.894,1,0.894}}0.9991                                 \\ 
\hline\hline
                       & \multicolumn{3}{c}{{\cellcolor[rgb]{0.918,0.918,0.918}}\textbf{Weight Merging (Evollm-jp-7b) }}                                                                                                                         & \multicolumn{3}{c}{{\cellcolor[rgb]{0.918,0.918,0.918}}\textbf{Distribution Merging(Fusellm-7b) }}                                                                                                                     \\
                       & Shisa-gamma-7b-v1                                                       & Wizardmath-7b-1.1                                                     & Abel-7b-002                                                           & Llama-2-7b                                                              & Openllama-2-7b                                                      & Mpt-7b                                                                 \\ 
\hline
\textbf{PCS}           & {\cellcolor[rgb]{0.894,1,0.894}}0.9992                                  & {\cellcolor[rgb]{0.894,1,0.894}}0.9990                                & {\cellcolor[rgb]{0.894,1,0.894}}0.9989                                & {\cellcolor[rgb]{0.894,1,0.894}}0.9997                                  & {\cellcolor[rgb]{1,0.894,0.894}}0.0194                              & {\cellcolor[rgb]{1,0.894,0.894}}0.0000                                 \\
\textbf{ICS}           & {\cellcolor[rgb]{0.894,1,0.894}}0.9992                                  & {\cellcolor[rgb]{0.894,1,0.894}}0.9988                                & {\cellcolor[rgb]{0.894,1,0.894}}0.9988                                & {\cellcolor[rgb]{1,0.894,0.894}}0.1043                                  & {\cellcolor[rgb]{1,0.894,0.894}}0.2478                              & {\cellcolor[rgb]{1,0.894,0.894}}0.1014                                 \\
\textbf{Logits}        & {\cellcolor[rgb]{0.894,1,0.894}}0.9933                                  & {\cellcolor[rgb]{0.894,1,0.894}}0.9999                                & {\cellcolor[rgb]{0.894,1,0.894}}0.9999                                & {\cellcolor[rgb]{0.894,1,0.894}}0.9999                                  & {\cellcolor[rgb]{1,0.894,0.894}}0.0100                              & {\cellcolor[rgb]{1,0.894,0.894}}0.0000                                 \\
\textbf{REEF}          & {\cellcolor[rgb]{0.894,1,0.894}}0.9635                                  & {\cellcolor[rgb]{0.894,1,0.894}}0.9526                                & {\cellcolor[rgb]{0.894,1,0.894}}0.9374                                & {\cellcolor[rgb]{0.894,1,0.894}}0.9996                                  & {\cellcolor[rgb]{0.894,0.894,1}}0.6713                              & {\cellcolor[rgb]{0.894,0.894,1}}0.6200                                 \\ 
\hline\hline
                       & \multicolumn{3}{c}{{\cellcolor[rgb]{0.918,0.918,0.918}}\textbf{Permutation }}                                                                                                                                           & \multicolumn{3}{c}{{\cellcolor[rgb]{0.918,0.918,0.918}}\textbf{Scaling Transformation}}                                                                                                                                \\
                       & Llama-2-7b                                                              & Mistral-7b                                                            & Qwen-1.5-7b                                                           & Llama-2-7b                                                              & Mistral-7b                                                          & Qwen-1.5-7b                                                            \\ 
\hline
\textbf{PCS}           & {\cellcolor[rgb]{1,0.894,0.894}}0.0000                                  & {\cellcolor[rgb]{1,0.894,0.894}}0.0000                                & {\cellcolor[rgb]{1,0.894,0.894}}0.0000                                & {\cellcolor[rgb]{0.894,1,0.894}}0.9999                                  & {\cellcolor[rgb]{0.894,1,0.894}}0.9989                              & {\cellcolor[rgb]{0.894,1,0.894}}0.9999                                 \\
\textbf{ICS}           & {\cellcolor[rgb]{1,0.894,0.894}}0.1918                                  & {\cellcolor[rgb]{0.894,1,0.894}}0.9847                                & {\cellcolor[rgb]{0.894,1,0.894}}0.9912                                & {\cellcolor[rgb]{0.894,1,0.894}}0.9999                                  & {\cellcolor[rgb]{0.894,1,0.894}}0.9999                              & {\cellcolor[rgb]{0.894,1,0.894}}0.9998                                 \\
\textbf{Logits}        & {\cellcolor[rgb]{1,0.894,0.894}}0.0000                                  & {\cellcolor[rgb]{1,0.894,0.894}}0.0000                                & {\cellcolor[rgb]{1,0.894,0.894}}0.0000                                & {\cellcolor[rgb]{0.894,1,0.894}}0.9999                                  & {\cellcolor[rgb]{0.894,1,0.894}}0.9999                              & {\cellcolor[rgb]{0.894,1,0.894}}0.9999                                 \\
\textbf{REEF}          & {\cellcolor[rgb]{0.894,1,0.894}}1.0000                                  & {\cellcolor[rgb]{0.894,1,0.894}}1.0000                               & {\cellcolor[rgb]{0.894,1,0.894}}1.0000                               & {\cellcolor[rgb]{0.894,1,0.894}}1.0000                                  & {\cellcolor[rgb]{0.894,1,0.894}}1.0000                              & {\cellcolor[rgb]{0.894,1,0.894}}1.0000                                \\
\hline\hline
\end{tabular}
}
\end{table}

%% file: section/5_conclusion.tex
\section{Conclusion}
\label{sec:conclusion}

This paper proposes REEF, a robust representation-based fingerprinting method for LLMs, which effectively identifies models derived from victim models. REEF does not impair LLMS's general capability and remains resilient against various subsequent developments, including pruning, fine-tuning, merging, and permutations. Therefore, REEF is highly suitable for protecting model IPs for both third parties and model owners, as a reliable solution for safeguarding models from unauthorized use or reproduction.

%% file: section/appendix.tex
\setcounter{equation}{0}
\setcounter{lemma}{0}
\setcounter{theorem}{0}
\setcounter{assumption}{0}

\section{Proof for Theorem \ref{theo:cka}}
\label{ap:proof}

\begin{theorem}
\label{theo:cka}
Given two matrices \( X \in \mathbb{R}^{m \times p_1} \) and \( Y \in \mathbb{R}^{m \times p_2} \), the CKA similarity score between \( X \) and \( Y \) is invariant under any permutation of the columns and column-wise scaling transformation. Formally, we have:
\begin{equation}
\label{eqn:cka_invariance}
\text{CKA}(X, Y) = \text{CKA}(XP_1, YP_2) = \text{CKA}(c_1X, c_2Y)
\end{equation}
where \( P_1 \in \mathbb{R}^{p_1 \times p_1} \) and \( P_2 \in \mathbb{R}^{p_2 \times p_2} \) denote permutation matrices. \( c_1 \in \mathbb{R}^{+} \) and \( c_2 \in \mathbb{R}^{+}\) are two positive scalars.
\end{theorem}

\textit{Proof.} \subsection{Case 1: Permutation Invariance}
\textbf{For Linear CKA}, the Gram matrices of $X$ and $Y$ are $ K_X = XX^{\top} $ and $K_Y = YY^{\top}$, respectively.

In this way, we have 

\begin{align}
    K_{XP_1} = (XP_1)(XP_1)^{\top}=X\underbrace{P_1P_1^{\top}}_{=I}X^{\top} = XX^{\top} = K_X.
\end{align}

  Since $P_1$ is an orthogonal permutation matrix, thus $P_1P_1^{\top} = I$.

Similarly, we have
\begin{align}
    K_{YP_2} = (YP_2)(YP_2)^{\top}= Y\underbrace{P_2P_2^{\top}}_{=I}Y^{\top} = YY^{\top} = K_Y.
\end{align}

According to \citep{gretton2005measuring},

\begin{align}
    \text{HSIC}(X, Y) &= \frac{1}{(m-1)^2}\text{tr}(K_XHK_YH) \nonumber\\
    &= \underbrace{\frac{1}{(m-1)^2}\text{tr}(K_{XP_1}HK_YH)}_{\text{HSIC}(XP_1, Y)} \nonumber\\
    &= \underbrace{\frac{1}{(m-1)^2}\text{tr}(K_{X}HK_{YP_2}H)}_{\text{HSIC}(X, YP_2)} \nonumber \\
    & = \underbrace{\frac{1}{(m-1)^2}\text{tr}(K_{XP1}HK_{YP_2}H)}_{\text{HSIC}(XP_1, YP_2)}
\end{align}

Thus, we have
\begin{equation}
\text{HSIC}(X, Y) = \text{HSIC}(XP_1, Y) = \text{HSIC}(X, YP_2) = \text{HSIC}(XP_1, YP_2).
\label{hsic}
\end{equation}

Taking Eq.\ref{hsic} into Eq. \ref{cka}, we have
\begin{align}
        \text{CKA}(X, Y) & = \frac{\text{HSIC}(X, Y)}{\sqrt{\text{HSIC}(X, X) \cdot \text{HSIC}(Y, Y)}} \nonumber \\
        &= \underbrace{\frac{\text{HSIC}(XP_1, Y)}{\sqrt{\text{HSIC}(XP_1, XP_1) \cdot \text{HSIC}(Y, Y)}}}_{\text{CKA}(XP_1, Y) } \nonumber \\
        &= \underbrace{\frac{\text{HSIC}(X, YP_2)}{\sqrt{\text{HSIC}(X, X) \cdot \text{HSIC}(YP_2, YP_2)}}}_{\text{CKA}(X, YP_2) } \nonumber \\
        & = \underbrace{\frac{\text{HSIC}(XP_1, YP_2)}{\sqrt{\text{HSIC}(XP_1, XP_1) \cdot \text{HSIC}(YP_2, YP_2)}}}_{\text{CKA}(XP_1, YP_2) }
\end{align}
Finally, we obtain 
\begin{equation}
\text{CKA}(X, Y) = \text{CKA}(XP_1, Y) = \text{CKA}(X, YP_2) = \text{CKA}(XP_1, YP_2)
\end{equation}

\textbf{For RBF CKA}, the RBF kernel function is 
\begin{align}
    k(X_i, X_j) & = \exp\left(-\frac{\|X_i - X_j\|_2^2}{2\sigma^2}\right) \nonumber \\
    & = \underbrace{\exp\left(-\frac{\|X_iP_1 - X_jP_1\|_2^2}{2\sigma^2}\right)}_{K(X_iP_1, X_jP_1)}
\end{align}

The pairwise distances $\|X_i - X_j\|_2$ are invariant to the column permutation of $X$, because $P_1$ is a permutation matrix. Therefore, we can obtain \(K_{XP_1} = K_X\).

Similarly, it is easily derived \(K_{YP_2} = K_Y\) as follows,
\begin{align}
    k(Y_i, Y_j) & = \exp\left(-\frac{\|Y_i - Y_j\|_2^2}{2\sigma^2}\right) \nonumber \\
    & = \underbrace{\exp\left(-\frac{\|Y_iP_2 - Y_jP_2\|_2^2}{2\sigma^2}\right)}_{K(Y_iP_2, Y_jP_2)}
\end{align}

In this way, we have 

\begin{align}
    \text{HSIC}(X, Y) &= \frac{1}{(m-1)^2}\text{tr}(K_XHK_YH) \nonumber\\
    &= \underbrace{\frac{1}{(m-1)^2}\text{tr}(K_{XP_1}HK_YH)}_{\text{HSIC}(XP_1, Y)} \nonumber\\
    &= \underbrace{\frac{1}{(m-1)^2}\text{tr}(K_{X}HK_{YP_2}H)}_{\text{HSIC}(X, YP_2)} \nonumber \\
    & = \underbrace{\frac{1}{(m-1)^2}\text{tr}(K_{XP_1}HK_{YP_2}H)}_{\text{HSIC}(XP_1, YP_2)}
    \label{hsic1}
\end{align}

Taking Eq.\ref{hsic1} into Eq. \ref{cka}, we have

\begin{align}
        \text{CKA}(X, Y) & = \frac{\text{HSIC}(X, Y)}{\sqrt{\text{HSIC}(X, X) \cdot \text{HSIC}(Y, Y)}} \nonumber \\
        &= \underbrace{\frac{\text{HSIC}(XP_1, Y)}{\sqrt{\text{HSIC}(XP_1, XP_1) \cdot \text{HSIC}(Y, Y)}}}_{\text{CKA}(XP_1, Y) } \nonumber \\
        &= \underbrace{\frac{\text{HSIC}(X, YP_2)}{\sqrt{\text{HSIC}(X, X) \cdot \text{HSIC}(YP_2, YP_2)}}}_{\text{CKA}(X, YP_2) } \nonumber \\
         & = \underbrace{\frac{\text{HSIC}(XP_1, YP_2)}{\sqrt{\text{HSIC}(XP_1, XP_1) \cdot \text{HSIC}(YP_2, YP_2)}}}_{\text{CKA}(XP_1, YP_2) }
\end{align}

Finally, we obtain 
\begin{equation}
\text{CKA}(X, Y) = \text{CKA}(XP_1, Y) = \text{CKA}(X, YP_2) = \text{CKA}(XP_1, YP_2) 
\end{equation}

\subsection{Case 2: Scaling Invariance}

\textbf{For Linear CKA}, let \( \tilde{X} = c_1X \) and $c_1 \in \mathbb{R}^{+}$. Then,
\begin{align}
  K_{\tilde{X}} & = \tilde{X}\tilde{X}^{\top} \nonumber \\
  & = (c_1X)(c_1X)^{\top} \nonumber \\
  & = c_1^2XX^{\top} \nonumber \\
  & = c_1^2K_X
\end{align}

Similarly, let \( \tilde{Y} = c_2Y \) and $c_2 \in \mathbb{R}^{+}$. Then,
\begin{align}
  K_{\tilde{Y}} & = \tilde{Y}\tilde{Y}^{\top} \nonumber \\
  & = (c_2Y)(c_2Y)^{\top} \nonumber \\
  & = c_2^2YY^{\top} \nonumber \\
  & = c_2^2K_Y.
\end{align}

In this way, 
\begin{align}
   \text{HSIC}(c_1X, c_2Y) & = \frac{1}{(m-1)^2} \text{tr}(K_{\tilde{X}} H K_{\tilde{Y}} H) \nonumber \\ 
   & = \frac{1}{(m-1)^2} \text{tr}(c_1^2K_X Hc_2^2 K_Y H) \nonumber \\
   & = \frac{1}{(m-1)^2} \text{tr}(c_1^2c_2^2K_X HK_Y H) \nonumber \\
   & = \frac{c_1^2c_2^2}{(m-1)^2} \text{tr}(K_X HK_Y H) \nonumber \\
   & = c_1^2c_2^2\text{HSIC}(X, Y).
\end{align}

Accordingly, 

\begin{align}
   \text{HSIC}(c_1X, c_1X) & = \frac{1}{(m-1)^2} \text{tr}(K_{\tilde{X}} H K_{\tilde{X}} H) \nonumber \\ 
   & = \frac{1}{(m-1)^2} \text{tr}(c_1^2K_X Hc_1^2 K_X H) \nonumber \\
   & = \frac{1}{(m-1)^2} \text{tr}(c_1^4K_X HK_X H) \nonumber \\
   & = \frac{c_1^4}{(m-1)^2} \text{tr}(K_X HK_X H) \nonumber \\
   & = c_1^4\text{HSIC}(X, X).
\end{align}

\begin{align}
   \text{HSIC}(c_2Y, c_2Y) & = \frac{1}{(m-1)^2} \text{tr}(K_{\tilde{Y}} H K_{\tilde{Y}} H) \nonumber \\ 
   & = \frac{1}{(m-1)^2} \text{tr}(c_2^2K_Y Hc_2^2 K_Y H) \nonumber \\
   & = \frac{1}{(m-1)^2} \text{tr}(c_2^4K_Y HK_Y H) \nonumber \\
   & = \frac{c_2^4}{(m-1)^2} \text{tr}(K_Y HK_Y H) \nonumber \\
   & = c_2^4\text{HSIC}(Y, Y).
\end{align}

Therefore, we have

\begin{align}
        \text{CKA}(c_1X, c_2Y) & = \frac{\text{HSIC}(c_1X, c_2Y)}{\sqrt{\text{HSIC}(c_1X, c_1X) \cdot \text{HSIC}(c_2Y, c_2Y)}} \nonumber \\
        &= \frac{c_1^2c_2^2\text{HSIC}(X, Y)}{\sqrt{c_1^4\text{HSIC}(X, X) \cdot c_2^4\text{HSIC}(Y, Y)}} \nonumber \\
        &= \frac{c_1^2c_2^2\text{HSIC}(X, Y)}{c_1^2c_2^2\sqrt{\text{HSIC}(X, X) \cdot \text{HSIC}(Y, Y)}} \nonumber \\
        &= \underbrace{\frac{\text{HSIC}(X, Y)}{\sqrt{\text{HSIC}(X, X) \cdot \text{HSIC}(Y, Y)}}}_{\text{CKA}(X, Y) }
\end{align}

Finally, we obtain 
\begin{equation}
\text{CKA}(X, Y)  = \text{CKA}(c_1X, c_2Y)
\end{equation}

\textbf{For RBF CKA}, the RBF kernel function is 
\begin{align}
    k(c_1X_i, c_1X_j) & = \exp\left(-\frac{\|c_1X_i - c_1X_j\|_2^2}{2\sigma^2}\right) \nonumber \\
    & = \exp\left(-\frac{{c_1^2}\|X_i - X_j\|_2^2}{2\sigma^2}\right)
    \label{rbf}
\end{align}

Following \cite{Kornblith2019}, the bandwidth \(\sigma\) is chosen as a fraction of the median distance, \ie, \(\sigma = \alpha \cdot \text{median}(\|X_i - X_j\|_2)\) for the constant \(\alpha > 0\).
In this way, Eq. \ref{rbf} is transformed as, 

\begin{align}
k(c_1X_i, c_1X_j) & = \exp\left(-\frac{{c_1^2}\|X_i - X_j\|_2^2}{2(\alpha c_1^2 \cdot \text{median}(\|X_i - X_j\|_2))^2}\right) \nonumber \\
& = \underbrace{\exp\left(-\frac{{c_1^2}\|X_i - X_j\|_2^2}{2c_1^2\sigma^2}\right)}_{k(X_i, X_j)}.
\end{align}

Similarly, it is easily derived $k(c_2Y_i, c_2Y_j) = k(Y_i,Y_j)$ as follows,
\begin{align}
k(c_2Y_i, c_2Y_j) & = \exp\left(-\frac{{c_2^2}\|Y_i - Y_j\|_2^2}{2(\alpha c_2^2 \cdot \text{median}(\|Y_i - Y_j\|_2))^2}\right) \nonumber \\
& = \underbrace{\exp\left(-\frac{{c_2^2}\|Y_i - Y_j\|_2^2}{2c_2^2\sigma^2}\right)}_{k(Y_i, Y_j)}.
\end{align}

Therefore, we can obtain $\text{HSIC}(X, Y) = \text{HSIC}(c_1X, c_2Y)$,  $\text{HSIC}(X, X) = \text{HSIC}(c_1X, c_1X)$, and $\text{HSIC}(Y, Y) = \text{HSIC}(c_2Y, c_2Y)$

Finally, we have 

\begin{align}
        \text{CKA}(c_1X, c_2Y) & = \frac{\text{HSIC}(c_1X, c_2Y)}{\sqrt{\text{HSIC}(c_1X, c_1X) \cdot \text{HSIC}(c_2Y, c_2Y)}} \nonumber \\
        &= \frac{\text{HSIC}(X, Y)}{\sqrt{\text{HSIC}(X, X) \cdot \text{HSIC}(Y, Y)}}  \nonumber \\
        &= \text{CKA}(X, Y).
\end{align}

Finally, we obtain 
\begin{equation}
\text{CKA}(X, Y) = \text{CKA}(c_1X, c_2Y).
\end{equation}

\section{The Effectiveness of Classifiers Trained on Representations of a Victim Model}
\label{sec:dnn-appx}

This appendix provides a detailed analysis of the experiments conducted to evaluate the effectiveness of classifiers trained on the representations of a victim model to identify whether a suspect model is derived from it, thus protecting its intellectual property. We explore the classifiers' accuracy when utilizing representations from different layers to train classifiers and applying them to the corresponding layers of the suspect model (\ref{subsec:same-layer}), as well as applying classifiers trained on one layer's representation to representations from other layers of the suspect model (\ref{subsec:diff-layer}).

\subsection{Apply Classifiers to The Corresponding Layer}
\label{subsec:same-layer}

Research has shown that representations from the middle and later layers of LLMs contain rich encoded information, which can be used to classify high-dimensional concepts, such as safety or unsafety, and honesty or dishonesty \citep{burns2022discovering,rimsky2023steering,zou2023representation,qian2024towards}. Following Section~\ref{sec:dnn}, we explore the effectiveness of classifiers trained on representations from different layers.

Specifically, we use Llama-2-7b and llama-2-13b as victim models, extracting representations from the 24th and 30th layers of Llama-2-7b and from the 32nd and 40th layers of Llama-2-13b for the TruthfulQA dataset. We then train various classifiers (\textit{e.g.}, linear, MLP, CNN, GCN) on representations from each layer. These classifiers are subsequently applied to various suspect models, including LLMs derived from the victim models as well as unrelated LLMs.

\begin{figure}[t]
    \centering
    \includegraphics[width=\textwidth]{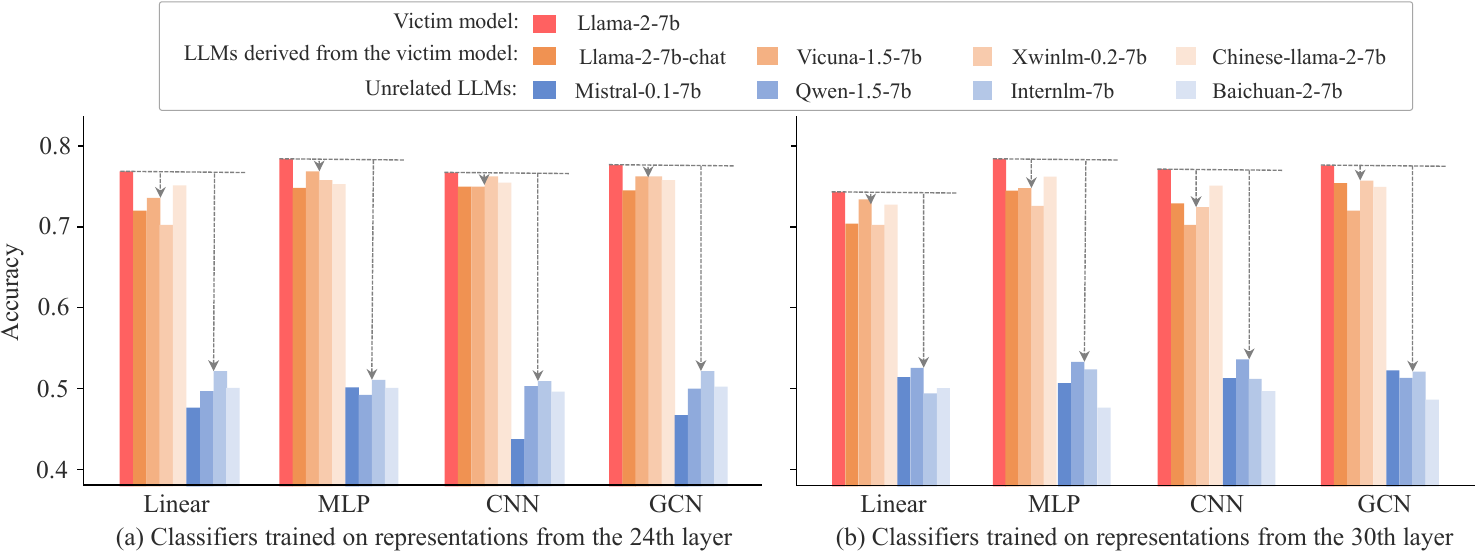} 
    \caption{Accuracies of classifiers trained on representations from Llama-2-7b.}
    \label{fig:same-layer-7b}
\end{figure}

\begin{figure}[t]
    \centering
    \includegraphics[width=\textwidth]{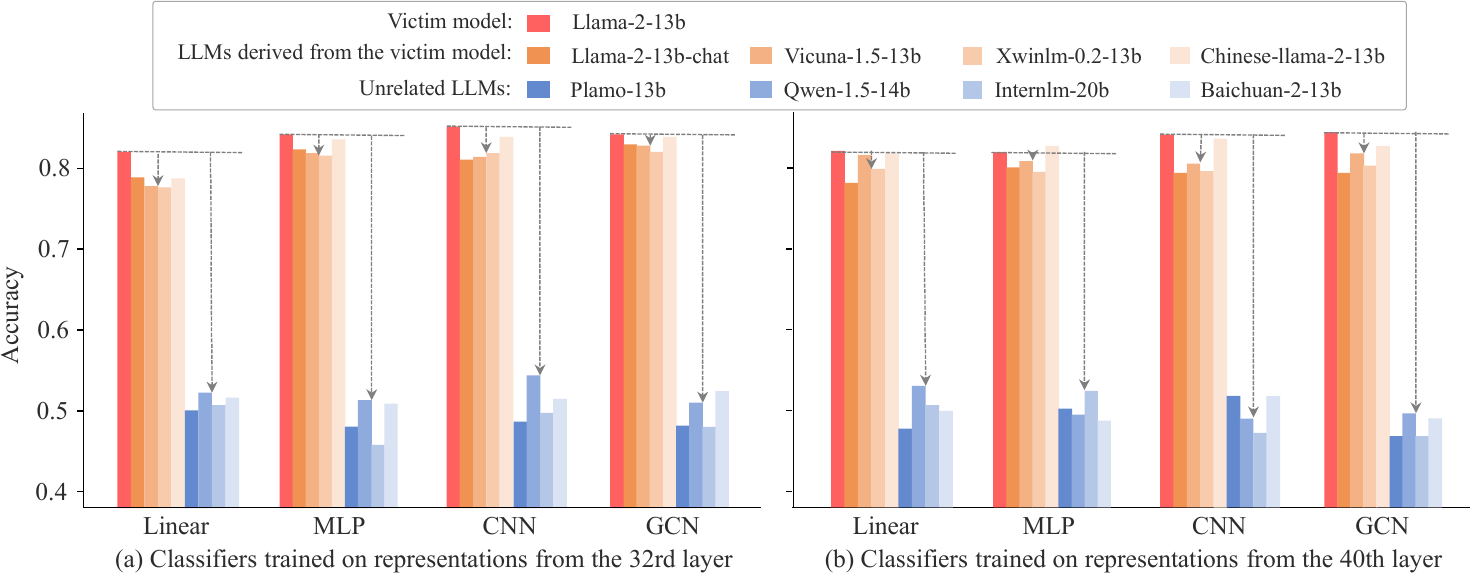} 
    \caption{Accuracies of classifiers trained on representations from Llama-2-13b.}
    \label{fig:same-layer-13b}
\end{figure}

Classifiers trained on representations from different layers of the victim model are all capable of identifying whether a suspect model is derived from the victim model. Figures \ref{fig:same-layer-7b} and \ref{fig:same-layer-13b} show the results of applying classifiers trained on representations from the 24th and 30th layers of Llama-2-7b and from the 32nd and 40th layers of Llama-2-13b to suspect models on the TruthfulQA dataset.
It can be observed that across different layers, all classifiers (linear, MLP, CNN, GCN) achieve an accuracy over 70\% on representations from LLMs derived from the victim model. This accuracy is close to the classification results of the victim model itself. However, the accuracy dropped to about 50\% when applied to representations from unrelated models, which is close to random guessing and significantly lower than the classification results on the victim model's representations.

The results demonstrate that REEF, our representation-based fingerprinting method, does not depend on representations from any specific layer. By leveraging the powerful representation modeling capabilities of LLMs, REEF can use representations from various layers to identify the victim model within a suspect model, thereby protecting its intellectual property.

\subsection{Apply Classifiers cross Layers}
\label{subsec:diff-layer}

To further investigate the generalizability of our approach, we conduct cross-layer experiments by applying classifiers trained on representations from one layer to representations from other layers. For instance, we apply a linear classifier trained on the 18th layer representations of Llama-2-7b to the 24th layer representations of suspect models. This cross-layer analysis provides insights into the similarity of representations across different layers of the model.

Following the same training process as previously described, for Llama-2-7b, we select one layer's representations from the 18th, 24th, or 30th layer to train a linear classifier, which is then applied to the representations from the other two layers across various suspect models. For instance, linear classifiers trained on representations from the 18th layer are applied to the representations of the 24th and 30th layers in different suspect models. Similarly, for Llama-2-13b, we choose representations from the 24th, 32nd, or 40th layer to conduct the same cross-layer classifier application. The experimental results are presented in Tables \ref{tab:diff-layer-7b} and \ref{tab:diff-layer-13b}, respectively, which provide detailed accuracy metrics for each cross-layer classification task.

Table \ref{tab:diff-layer-7b} shows that the classifier trained on the specific layer's representations (\eg, 18th layer) of Llama-2-7b, when applied to other layers' representations (\eg, 24th and 30th layer) of suspect models, maintained the accuracy 70\% for derived models and 50\% for unrelated models. Table \ref{tab:diff-layer-13b} demonstrates similar results for experiments conducted on the larger Llama-2-13b model, with significantly distinct accuracy ranges. These results indicate that classifiers trained on one layer's representations remain effective when applied to other layers, suggesting a significant similarity in the representation spaces across different layers of the model.

The ability of these classifiers to generalize across layers further strengthens the reliability of our fingerprinting detection method. It indicates that the distinctive features learned by the classifiers are not confined to a specific layer but are present throughout the model's architecture. This characteristic enhances the robustness of our approach, making the use of representations as fingerprints for protecting the intellectual property of the victim model more reliable through cross-layer validation. Additionally, this insight inspires us to use heatmaps to depict the CKA similarity between the representations of the victim LLM and those of various suspect LLMs across the same samples, as presented in the main text.

\begin{table}[t]
\centering
\setlength{\extrarowheight}{0pt}
\addtolength{\extrarowheight}{\aboverulesep}
\addtolength{\extrarowheight}{\belowrulesep}
\setlength{\aboverulesep}{0pt}
\setlength{\belowrulesep}{0pt}
\caption{Accuracies of classifiers applied across layers for victim model Llama-2-7b. \colorbox[rgb]{0.945,0.945,0.945}{Gray shading} indicates that the classifier was trained using representations from that specific layer.}
\label{tab:diff-layer-7b}
\scalebox{0.8}{
\begin{tabular}{cccccccccc} 
\toprule
        & \textbf{Victim LLM}
        & \multicolumn{4}{c}{\textbf{LLMs derived from the victim model}}
        & \multicolumn{4}{c}{\textbf{Unrelated LLMs}}
        \\ 
\cdashline{2-10}
        & \begin{tabular}[c]{@{}c@{}}Llama-2\\-7b\end{tabular} & \begin{tabular}[c]{@{}c@{}}Llama-2\\-7b-chat\end{tabular} & \begin{tabular}[c]{@{}c@{}}Vicuna-1.5\\-7b\end{tabular} & \begin{tabular}[c]{@{}c@{}}Chinese-\\llama-2-7b\end{tabular} & \begin{tabular}[c]{@{}c@{}}Xwimlm\\-7b\end{tabular} & \begin{tabular}[c]{@{}c@{}}Mistral\\-7b\end{tabular} & \begin{tabular}[c]{@{}c@{}}Baichuan\\-2-7b\end{tabular} & \begin{tabular}[c]{@{}c@{}}Qwen\\-1.5-7b\end{tabular} & \begin{tabular}[c]{@{}c@{}}Internlm\\-7b\end{tabular}  \\ 
\hline
\rowcolor[rgb]{0.945,0.945,0.945} \textbf{Layer-18}          & \textcolor[rgb]{0,0.106,0.627}{0.8003}               & \textcolor[rgb]{0,0.106,0.627}{0.7437}                    & \textcolor[rgb]{0,0.106,0.627}{0.7642}                  & \textcolor[rgb]{0,0.106,0.627}{0.7578}                       & \textcolor[rgb]{0,0.106,0.627}{0.7421}              & \textcolor[rgb]{0.769,0.416,0.463}{0.5078}           & \textcolor[rgb]{0.769,0.416,0.463}{0.4513}              & \textcolor[rgb]{0.769,0.416,0.463}{0.5063}            & \textcolor[rgb]{0.769,0.416,0.463}{0.5094}             \\
\textbf{\textbf{L}ayer-24}                                   & \textcolor[rgb]{0,0.106,0.627}{0.7123}               & \textcolor[rgb]{0,0.106,0.627}{0.7008}                    & \textcolor[rgb]{0,0.106,0.627}{0.6965}                  & \textcolor[rgb]{0,0.106,0.627}{0.7081}                       & \textcolor[rgb]{0,0.106,0.627}{0.7060}              & \textcolor[rgb]{0.769,0.416,0.463}{0.4953}           & \textcolor[rgb]{0.769,0.416,0.463}{0.5314}              & \textcolor[rgb]{0.769,0.416,0.463}{0.5283}            & \textcolor[rgb]{0.769,0.416,0.463}{0.5016}             \\
\textbf{Layer-30}                                   & \textcolor[rgb]{0,0.106,0.627}{0.6715}               & \textcolor[rgb]{0,0.106,0.627}{0.6778}                    & \textcolor[rgb]{0,0.106,0.627}{0.6809}                  & \textcolor[rgb]{0,0.106,0.627}{0.6762}                       & \textcolor[rgb]{0,0.106,0.627}{0.6636}              & \textcolor[rgb]{0.769,0.416,0.463}{0.5031}           & \textcolor[rgb]{0.769,0.416,0.463}{0.4890}              & \textcolor[rgb]{0.769,0.416,0.463}{0.5094}            & \textcolor[rgb]{0.769,0.416,0.463}{0.5252}             \\ 
\midrule
\textbf{\textbf{L}ayer-18}                                   & \textcolor[rgb]{0,0.106,0.627}{0.7014}               & \textcolor[rgb]{0,0.106,0.627}{0.7030}                    & \textcolor[rgb]{0,0.106,0.627}{0.7124}                  & \textcolor[rgb]{0,0.106,0.627}{0.7077}                       & \textcolor[rgb]{0,0.106,0.627}{0.6967}              & \textcolor[rgb]{0.769,0.416,0.463}{0.4717}           & \textcolor[rgb]{0.769,0.416,0.463}{0.5283}              & \textcolor[rgb]{0.769,0.416,0.463}{0.5418}            & \textcolor[rgb]{0.769,0.416,0.463}{0.5130}             \\
\rowcolor[rgb]{0.945,0.945,0.945} \textbf{\textbf{L}ayer-24} & \textcolor[rgb]{0,0.106,0.627}{0.7720}               & \textcolor[rgb]{0,0.106,0.627}{0.7233}                    & \textcolor[rgb]{0,0.106,0.627}{0.7390}                  & \textcolor[rgb]{0,0.106,0.627}{0.7055}                       & \textcolor[rgb]{0,0.106,0.627}{0.7547}              & \textcolor[rgb]{0.769,0.416,0.463}{0.4780}           & \textcolor[rgb]{0.769,0.416,0.463}{0.4984}              & \textcolor[rgb]{0.769,0.416,0.463}{0.5235}            & \textcolor[rgb]{0.769,0.416,0.463}{0.5031}             \\
\textbf{Layer-30}                                   & \textcolor[rgb]{0,0.106,0.627}{0.6723}               & \textcolor[rgb]{0,0.106,0.627}{0.6629}                    & \textcolor[rgb]{0,0.106,0.627}{0.7085}                  & \textcolor[rgb]{0,0.106,0.627}{0.6660}                       & \textcolor[rgb]{0,0.106,0.627}{0.6975}              & \textcolor[rgb]{0.769,0.416,0.463}{0.4513}           & \textcolor[rgb]{0.769,0.416,0.463}{0.4953}              & \textcolor[rgb]{0.769,0.416,0.463}{0.5126}            & \textcolor[rgb]{0.769,0.416,0.463}{0.4764}             \\ 
\midrule
\textbf{Layer-18}                                   & \textcolor[rgb]{0,0.106,0.627}{0.6982}               & \textcolor[rgb]{0,0.106,0.627}{0.6945}                    & \textcolor[rgb]{0,0.106,0.627}{0.6914}                  & \textcolor[rgb]{0,0.106,0.627}{0.6950}                       & \textcolor[rgb]{0,0.106,0.627}{0.6840}              & \textcolor[rgb]{0.769,0.416,0.463}{0.5225}           & \textcolor[rgb]{0.769,0.416,0.463}{0.5096}              & \textcolor[rgb]{0.769,0.416,0.463}{0.4827}            & \textcolor[rgb]{0.769,0.416,0.463}{0.5189}             \\
\textbf{Layer-24}                                   & \textcolor[rgb]{0,0.106,0.627}{0.7097}               & \textcolor[rgb]{0,0.106,0.627}{0.7050}                    & \textcolor[rgb]{0,0.106,0.627}{0.7191}                  & \textcolor[rgb]{0,0.106,0.627}{0.7034}                       & \textcolor[rgb]{0,0.106,0.627}{0.7233}              & \textcolor[rgb]{0.769,0.416,0.463}{0.5189}           & \textcolor[rgb]{0.769,0.416,0.463}{0.4959}              & \textcolor[rgb]{0.769,0.416,0.463}{0.4591}            & \textcolor[rgb]{0.769,0.416,0.463}{0.4686}             \\
\rowcolor[rgb]{0.945,0.945,0.945} \textbf{Layer-30} & \textcolor[rgb]{0,0.106,0.627}{0.7453}               & \textcolor[rgb]{0,0.106,0.627}{0.7061}                    & \textcolor[rgb]{0,0.106,0.627}{0.7360}                  & \textcolor[rgb]{0,0.106,0.627}{0.7045}                       & \textcolor[rgb]{0,0.106,0.627}{0.7296}              & \textcolor[rgb]{0.769,0.416,0.463}{0.5157}           & \textcolor[rgb]{0.769,0.416,0.463}{0.5270}              & \textcolor[rgb]{0.769,0.416,0.463}{0.4953}            & \textcolor[rgb]{0.769,0.416,0.463}{0.5036}             \\
\bottomrule
\end{tabular}
}
\end{table}

\begin{table}[t]
\centering
\setlength{\extrarowheight}{0pt}
\addtolength{\extrarowheight}{\aboverulesep}
\addtolength{\extrarowheight}{\belowrulesep}
\setlength{\aboverulesep}{0pt}
\setlength{\belowrulesep}{0pt}
\caption{Accuracies of classifiers applied across layers for victim model Llama-2-13b. \colorbox[rgb]{0.945,0.945,0.945}{Gray shading} indicates that the classifier was trained using representations from that specific layer.}
\label{tab:diff-layer-13b}
\scalebox{0.77}{
\begin{tabular}{cccccccccc} 
\toprule
        & \textbf{Victim model}
        & \multicolumn{4}{c}{\textbf{LLMs derived from the victim model }}
        & \multicolumn{4}{c}{\textbf{Unrelated LLMs }}
        \\ 
\cdashline{2-10}
        & \begin{tabular}[c]{@{}c@{}}Llama-2\\-13b\end{tabular} & \begin{tabular}[c]{@{}c@{}}Llama-2\\-13b-chat\end{tabular} & \begin{tabular}[c]{@{}c@{}}Vicuna-1.5\\-13b\end{tabular} & \begin{tabular}[c]{@{}c@{}}Chinese-\\llama-2-13b\end{tabular} & \begin{tabular}[c]{@{}c@{}}Xwimlm\\-13b\end{tabular} & \begin{tabular}[c]{@{}c@{}}Plamo\\-13b\end{tabular} & \begin{tabular}[c]{@{}c@{}}Baichuan\\-2-13b\end{tabular} & \begin{tabular}[c]{@{}c@{}}Qwen\\-1.5-14b\end{tabular} & \begin{tabular}[c]{@{}c@{}}Internlm\\-20b\end{tabular}  \\ 
\midrule
\rowcolor[rgb]{0.945,0.945,0.945} \textbf{Layer-24} & \textcolor[rgb]{0,0.106,0.627}{0.8412}                & \textcolor[rgb]{0,0.106,0.627}{0.8223}                     & \textcolor[rgb]{0,0.106,0.627}{0.8066}                   & \textcolor[rgb]{0,0.106,0.627}{0.8081}                        & \textcolor[rgb]{0,0.106,0.627}{0.8223}               & \textcolor[rgb]{0.769,0.416,0.463}{0.4827}          & \textcolor[rgb]{0.769,0.416,0.463}{0.5283}               & \textcolor[rgb]{0.769,0.416,0.463}{0.4276}             & \textcolor[rgb]{0.769,0.416,0.463}{0.4946}              \\
\textbf{Layer-32}                                   & \textcolor[rgb]{0,0.106,0.627}{0.8050}                 & \textcolor[rgb]{0,0.106,0.627}{0.7783}                     & \textcolor[rgb]{0,0.106,0.627}{0.7814}                   & \textcolor[rgb]{0,0.106,0.627}{0.7909}                        & \textcolor[rgb]{0,0.106,0.627}{0.8082}               & \textcolor[rgb]{0.769,0.416,0.463}{0.4811}          & \textcolor[rgb]{0.769,0.416,0.463}{0.4827}               & \textcolor[rgb]{0.769,0.416,0.463}{0.4450}              & \textcolor[rgb]{0.769,0.416,0.463}{0.4546}              \\
\textbf{Layer-40}                                   & \textcolor[rgb]{0,0.106,0.627}{0.7767}                & \textcolor[rgb]{0,0.106,0.627}{0.7248}                     & \textcolor[rgb]{0,0.106,0.627}{0.7783}                   & \textcolor[rgb]{0,0.106,0.627}{0.7421}                        & \textcolor[rgb]{0,0.106,0.627}{0.7594}               & \textcolor[rgb]{0.769,0.416,0.463}{0.4780}           & \textcolor[rgb]{0.769,0.416,0.463}{0.5372}               & \textcolor[rgb]{0.769,0.416,0.463}{0.4906}             & \textcolor[rgb]{0.769,0.416,0.463}{0.4289}              \\ 
\midrule
\textbf{Layer-24}                                   & \textcolor[rgb]{0,0.106,0.627}{0.8381}                & \textcolor[rgb]{0,0.106,0.627}{0.7925}                     & \textcolor[rgb]{0,0.106,0.627}{0.8113}                   & \textcolor[rgb]{0,0.106,0.627}{0.8145}                        & \textcolor[rgb]{0,0.106,0.627}{0.8192}               & \textcolor[rgb]{0.769,0.416,0.463}{0.4874}          & \textcolor[rgb]{0.769,0.416,0.463}{0.5329}               & \textcolor[rgb]{0.769,0.416,0.463}{0.5236}             & \textcolor[rgb]{0.769,0.416,0.463}{0.4996}              \\
\rowcolor[rgb]{0.945,0.945,0.945} \textbf{Layer-32} & \textcolor[rgb]{0,0.106,0.627}{0.8223}                & \textcolor[rgb]{0,0.106,0.627}{0.7909}                     & \textcolor[rgb]{0,0.106,0.627}{0.7799}                   & \textcolor[rgb]{0,0.106,0.627}{0.7799}                        & \textcolor[rgb]{0,0.106,0.627}{0.7909}               & \textcolor[rgb]{0.769,0.416,0.463}{0.5000}             & \textcolor[rgb]{0.769,0.416,0.463}{0.5220}                & \textcolor[rgb]{0.769,0.416,0.463}{0.5079}             & \textcolor[rgb]{0.769,0.416,0.463}{0.5057}              \\
\textbf{Layer-40}                                   & \textcolor[rgb]{0,0.106,0.627}{0.7767}                & \textcolor[rgb]{0,0.106,0.627}{0.7484}                     & \textcolor[rgb]{0,0.106,0.627}{0.7642}                   & \textcolor[rgb]{0,0.106,0.627}{0.7186}                        & \textcolor[rgb]{0,0.106,0.627}{0.7767}               & \textcolor[rgb]{0.769,0.416,0.463}{0.5083}          & \textcolor[rgb]{0.769,0.416,0.463}{0.5152}               & \textcolor[rgb]{0.769,0.416,0.463}{0.5350}              & \textcolor[rgb]{0.769,0.416,0.463}{0.4893}              \\ 
\midrule
\textbf{Layer-24}                                   & \textcolor[rgb]{0,0.106,0.627}{0.8302}                & \textcolor[rgb]{0,0.106,0.627}{0.827}                      & \textcolor[rgb]{0,0.106,0.627}{0.8129}                   & \textcolor[rgb]{0,0.106,0.627}{0.8113}                        & \textcolor[rgb]{0,0.106,0.627}{0.8223}               & \textcolor[rgb]{0.769,0.416,0.463}{0.4858}          & \textcolor[rgb]{0.769,0.416,0.463}{0.5412}               & \textcolor[rgb]{0.769,0.416,0.463}{0.5000}                & \textcolor[rgb]{0.769,0.416,0.463}{0.4734}              \\
\textbf{Layer-32}                                   & \textcolor[rgb]{0,0.106,0.627}{0.8113}                & \textcolor[rgb]{0,0.106,0.627}{0.7783}                     & \textcolor[rgb]{0,0.106,0.627}{0.8035}                   & \textcolor[rgb]{0,0.106,0.627}{0.7814}                        & \textcolor[rgb]{0,0.106,0.627}{0.8003}               & \textcolor[rgb]{0.769,0.416,0.463}{0.4560}           & \textcolor[rgb]{0.769,0.416,0.463}{0.5397}               & \textcolor[rgb]{0.769,0.416,0.463}{0.5031}             & \textcolor[rgb]{0.769,0.416,0.463}{0.4896}              \\
\rowcolor[rgb]{0.945,0.945,0.945} \textbf{Layer-40} & \textcolor[rgb]{0,0.106,0.627}{0.8239}                & \textcolor[rgb]{0,0.106,0.627}{0.7842}                     & \textcolor[rgb]{0,0.106,0.627}{0.8187}                   & \textcolor[rgb]{0,0.106,0.627}{0.8014}                        & \textcolor[rgb]{0,0.106,0.627}{0.8207}               & \textcolor[rgb]{0.769,0.416,0.463}{0.4780}           & \textcolor[rgb]{0.769,0.416,0.463}{0.5314}               & \textcolor[rgb]{0.769,0.416,0.463}{0.5173}             & \textcolor[rgb]{0.769,0.416,0.463}{0.5000}                 \\
\bottomrule
\end{tabular}
}
\end{table}

\section{Heatmaps of the Victim Model and Different Suspect Models}
\label{sec:heatmap}

\begin{figure}[ht]
    \centering
    \includegraphics[width=\textwidth]{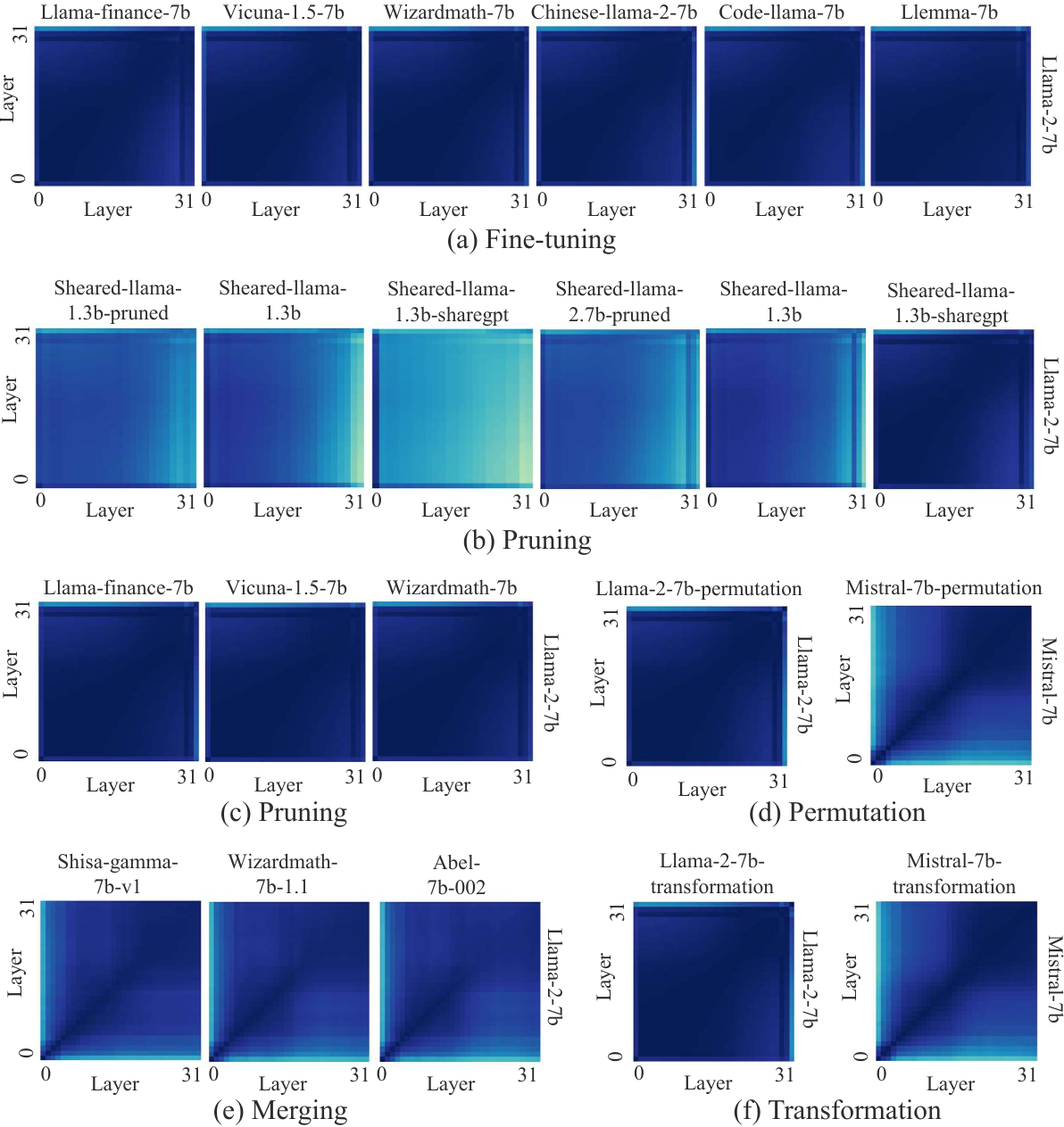} 
    \caption{Heatmaps depicting the layer-wise and inter-layer CKA similarity of representations for the same samples between each pair of victim and suspect models.}
    \label{fig:heatmap-robust}
\end{figure}

In Section~\ref{subsec:robust}, we report REEF's similarity of representations from the 18th layer between the victim model and various suspect models. These suspect models are derived from the victim model through a range of developments, including fine-tuning, pruning, merging, permutation, and scaling transformation. To provide a clearer and more intuitive comparison, we supplement this analysis with heatmaps in Figure~\ref{fig:heatmap-robust}, depicting the layer-wise and inter-layer CKA similarity of representations for the same samples between each pair of victim and suspect models. Figure~\ref{fig:heatmap-robust} demonstrates that, regardless of the type of development applied to the victim model, our representation-based fingerprint REEF can significantly identify the victim model, as shown by the high CKA similarities in the heatmap.

\section{Evading REEF with Fine-tuning}
\label{sec:fine-tuning}

We hypothesize that malicious developers aware of the REEF approach might attempt to design customized loss functions during fine-tuning to evade detection. Given that REEF determines model similarity based on the representation similarity between the suspect and victim models, malicious developers aiming to avoid detection would likely design their customized loss to maximize the representational divergence between these models.

Based on this premise, we designed two experiments to attempt to circumvent REEF detection:

\begin{itemize}[leftmargin=*]
    \item Integrating the task loss with a customized loss during the fine-tuning process, aiming to achieve the fine-tuning objective while maximizing the representational dissimilarity with the victim model.
    \item Fine-tuning the victim model solely using the customized loss, attempting to maximize the representational dissimilarity between the original and fine-tuned models.
\end{itemize}

To evaluate these scenarios, we conduct experiments using the OPT-1.3B model \citep{zhang2022opt} and the E2E NLG Challenge dataset \citep{novikova2017e2e} for fine-tuning. We employ the LoRA technique \citep{hu2021lora} for efficient adaptation. The customized loss is designed to measure the CKA similarity between the logits of the original and fine-tuned models.

For the first scenario, we formulate a combined loss function:
$\mathcal{L} = \mathcal{L}_{\text{task}} + \lambda \mathcal{L}_{\text{custom}}$,
where $\mathcal{L}_{\text{task}}$ is the task-specific loss (e.g., cross-entropy for the E2E NLG Challenge), $\mathcal{L}_{\text{custom}}$ is the CKA similarity between the logits of the original and fine-tuned models, and $\lambda$ is the weighting coefficient. Specifically, the customized loss is calculated using Equation \ref{cka}, that is:

\begin{align}
    \text{CKA}(\text{LG}_{\text{ori}}, \text{LG}_{\text{ft}}) = \frac{\text{HSIC}(\text{LG}_{\text{ori}}, \text{LG}_{\text{ft}})}{\sqrt{\text{HSIC}(\text{LG}_{\text{ori}}, \text{LG}_{\text{ori}}) \cdot \text{HSIC}(\text{LG}_{\text{ft}}, \text{LG}_{\text{ft}})}},
\end{align}

where $\text{LG}_{\text{ori}}$ and $\text{LG}_{\text{ft}}$ represent the logits of the original and fine-tuned models on the same sample.

In this scenario, incorporating different weighting coefficients ($\lambda$ ranges from 0.5 to 3.0) for the customized loss during the combined fine-tuning process failed to reduce the representational similarity between the fine-tuned model and the original model. This suggests that during fine-tuning, the model continues to rely on the representation modeling capabilities of the original language model. Consequently, achieving ECE task objectives necessarily preserves the representational distribution.

In the second scenario, although targeted fine-tuning can increase the distributional divergence in the representation space between the suspect and victim models, the suspect model loses its fundamental language expression capabilities, rendering its outputs meaningless. For example, the fine-tuned model may only respond with repetitive patterns such as ``and and and and ...'' for any input, demonstrating a complete loss of linguistic coherence and utility.

Therefore, our method demonstrates resilience against malicious actors' attempts to evade detection through fine-tuning strategies. These findings underscore the robustness of REEF in identifying the victim model, even in the face of sophisticated evasion techniques.

\section{Limitations}
\label{sec:limitations}

There are several limitations to this work.
Firstly, our study focuses on open-source LLMs, which allows model owners and third parties (e\textit{.g.}, regulatory authorities) to verify and protect model ownership. However, for closed-source models, the lack of access to their representations limits the applicability of our approach.
Secondly, regarding fine-tuning, due to the high cost of fine-tuning with extensive data (more than 700B), although we discuss the effectiveness of our method in main paper, empirical validation is lacking.